\documentclass[preprint,12pt]{elsarticle}

\usepackage{amssymb}

\usepackage{hyperref}
\usepackage{amsmath,amsfonts}
\usepackage{algorithmic}
\usepackage[ruled,linesnumbered]{algorithm2e}
\usepackage{array}
\usepackage[caption=false,font=normalsize,labelfont=sf,textfont=sf]{subfig}
\usepackage{textcomp}
\usepackage{stfloats}
\usepackage{url}
\usepackage{verbatim}
\usepackage{graphicx}

\usepackage{multirow}
\usepackage[capitalize]{cleveref} 
\crefname{section}{Sec.}{Secs.}
\Crefname{section}{Section}{Sections}
\Crefname{table}{Table}{Tables}
\crefname{table}{Tab.}{Tabs.}
\newcommand{\myPara}[1]{\vspace{.05in}\noindent\textbf{#1}}
\newcommand{\etal}[0]{\emph{et al.}}
\usepackage{tikz}
\newcommand*{\circled}[1]{\lower.7ex\hbox{\tikz\draw (0pt, 0pt)%
    circle (.5em) node {\makebox[1em][c]{\small #1}};}}
    
\journal{Neural Networks}

\begin{document}

\begin{frontmatter}



\title{MotionTrack: Learning Motion Predictor for Multiple Object Tracking}


\author[label1]{Changcheng Xiao}
\ead{xiaocc612@foxmail.com}

\author[label2]{Qiong Cao}

\author[label3]{Yujie Zhong}

\author[label5]{Long Lan}

\author[label4]{Xiang Zhang}

\author[label1]{Zhigang Luo}

\author[label2]{Dacheng Tao}

\affiliation[label1]{organization={School of Computer Science, National University of Defense Technology},
            city={Changsha},
            postcode={410073}, 
            state={Hunan},
            country={China}}
\affiliation[label2]{organization={JD Explore Academy},
            city={Beijing},
            postcode={102628}, 
            country={China}}
\affiliation[label3]{organization={Meituan Inc.},
            city={Beijing},
            postcode={100000}, 
            country={China}}
\affiliation[label4]{organization={Laboratory of Digitizing Software for Frontier Equipment, National University of Defense Technology},
            city={Changsha},
            postcode={410073}, 
            state={Hunan},
            country={China}}
\affiliation[label5]{organization={Institute for Quantum \& State Key Laboratory of High Performance Computing, National University of Defense Technology},
            city={Changsha},
            postcode={410073}, 
            state={Hunan},
            country={China}}

\begin{abstract}
Significant progress has been achieved in multi-object tracking (MOT) through the evolution of detection and re-identification (ReID) techniques. Despite these advancements, accurately tracking objects in scenarios with homogeneous appearance and heterogeneous motion remains a challenge. This challenge arises from two main factors: the insufficient discriminability of ReID features and the predominant utilization of linear motion models in MOT. In this context, we introduce a novel motion-based tracker, MotionTrack, centered around a learnable motion predictor that relies solely on object trajectory information. This predictor comprehensively integrates two levels of granularity in motion features to enhance the modeling of temporal dynamics and facilitate precise future motion prediction for individual objects.
Specifically, the proposed approach adopts a self-attention mechanism to capture token-level information and a Dynamic MLP layer to model channel-level features. MotionTrack is a simple, online tracking approach. Our experimental results demonstrate that MotionTrack yields state-of-the-art performance on datasets such as Dancetrack and SportsMOT, characterized by highly complex object motion.
\end{abstract}




\end{frontmatter}


\section{Introduction}
\label{sec:intro}
Multi-object tracking (MOT) has received more and more attention in recent years due to its promising applications in the fields 
\cite{wen2020ua,glamr,nuscenes,sun2020scalability,jrdb} of intelligent surveillance, autonomous driving, mobile robotics, etc. Benefiting from the rapid development of object detection \cite{fasterrcnn,yolov3,centernet,yolox,yang2019towards} and re-identification (ReID) \cite{deepsort,hermans2017defense,jde,fairmot}, tracking-by-detection methods \cite{wang2021two} have dominated. 
This paradigm consists of two main steps: 1) using an off-the-shelf object detector to obtain detection results for each frame, and 2) associating the detection results into trajectories using visual and motion cues. At association step, influenced by the inherent characteristics of the existing multi-object/pedestrian benchmarks \cite{mot16,mot20}, most recent successes on MOT \cite{jde,fairmot,trackformer} are based on either the appearance features or a junction of detection and simple tracking mechanism, leaving motion information under-explored. This trend makes existing trackers fail in situations \cite{SportsMOT, dancetrack} where objects of interest share very similar appearance in group dancing and players possess rapid motion in sports scenes. This observation motivates us to only incorporate motion cues into modelling, proving crucial for accurately and efficiently associating objects across frames in these complex situations.



In this work, we focus on learning a motion predictor to boost the accuracy of association and thus the  performance of tracking. It is very challenging due to complex motion variations across different scenarios and severe occlusions. The motion models used by exiting trackers can be divided into classical algorithms based on Bayesian estimation \cite{sort,deepsort,motdt,fairmot,bytetrack} and data-driven algorithms \cite{FFT,centertrack,artist,Tracktor}. For the former, a representative is Kalman filter\cite{kalman}. It assumes constant velocity and thus works well with linear motion, but tends to fail in handling nonlinear motion \cite{dancetrack,SportsMOT}. For the latter, the optical flow-based tracking method \cite{FFT} requires a complex and heavy optical flow model to calculate the inter-frame pixel offsets, which can only consider local motion information and is limited by the time-consuming optical flow model \cite{centertrack}. Furthermore, methods relying on Long Short-Term Memory (LSTM) networks \cite{artist,deft} have been utilized to store the motion information of objects within their hidden states. However, LSTMs have faced criticism for their memory mechanism \cite{luo2018fast} and their limited ability to model long-term temporal interactions \cite{bai2018empirical}. 
These step-by-step prediction methods lead to error accumulation\cite{ocsort,zhou2021informer}, which may result in inaccuracies when predicting the spatial locations of objects.


\begin{figure}[!t]
    \centering
    \subfloat[OC\_SORT]{\includegraphics[width=0.9\linewidth]{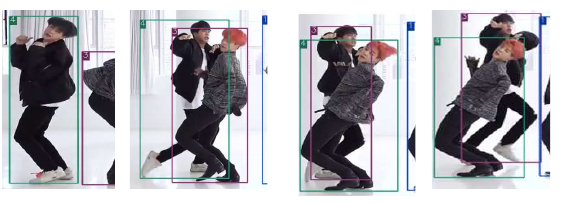}}
    \label{fig:short-b}
   \vfill
    \subfloat[MotionTrack]{\includegraphics[width=0.9\linewidth]{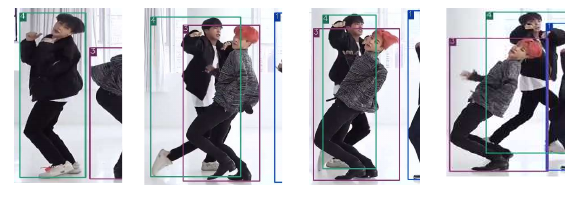}}
    \label{fig:short-c}
  \caption{A qualitative comparison between the proposed tracker and OC\_SORT is presented in a typical nonlinear motion scene. Samples were extracted from frames 87, 126, 128, and 132 of the video \textit{Dancetrack0058}. In the sequence, as the black-clad dancer turns around and crosses paths with the red-haired dancer, OC\_SORT experiences ID switches (4 $\xrightarrow{}$ 3), while our tracker successfully continues tracking.}
  \label{fig:comparision} 
\end{figure}

To address the aforementioned challenges and efficiently utilize object trajectory information, we propose a new online tracker, MotionTrack, based on a motion predictor that directly takes longer trajectories as inputs, predicting the future location of objects. As shown in \Cref{fig:comparision}, our proposed method can robustly track objects in scenarios with occlusion and nonlinear motion, while the Kalman filter-based trackers \cite{sort,ocsort} fail.
Specifically, we make the following contributions.


Firstly, in order to address the limitations of existing works \cite{sort, deepsort, ocsort, artist, deft}, which typically process observations sequentially before predicting bounding boxes of objects autoregressively, we explore to harness the powerful long-range dependency modeling capability of the Transformer \cite{transformer}. To maintain algorithmic simplicity and computational efficiency, we exclusively design a motion predictor based on the Transformer encoder to model the long-term trajectory information of individual objects for motion prediction. Specifically, the proposed motion predictor considers multiple observations simultaneously, weighting trajectory embeddings based on token-level pair-wise similarities \cite{transformer}.

Secondly, to further enhance the utilization of trajectory embedding sequences, we introduce a Multi-Layer Perceptron-like architecture named Dynamic MLP. As commonly recognized, different semantic information in the feature space tends to be distributed across distinct channels \cite{bau2020understanding, wu2021stylespace}, and channel mixing provides greater flexibility to explore cross-channel interaction \cite{Li2023MTSMixersMT, zhang2023crossformer, chen2023tsmixer}. Inspired by this, we aim to design a more powerful attention module to advance complex motion modeling and capture information distributed across different channels, such as relative position changes and directions of object motion. The proposed Dynamic MLP can precisely explore motion information distributed in different channels within the non-local range through content-adaptive token-mixing. With the Dynamic MLP, we further integrate it with the self-attention module in the Transformer \cite{transformer} to enable message passing at two different granularities, namely, token level and channel level, for the purpose of aggregating different semantic messages.

Moreover, we obtain more complex motion patterns via different data augmentations to better understand motion dynamics and boost the tracking performance. Specifically, random drop, random spatial jitter and random length are adopted to create motions with fast-moving objects, tracklets of different lengths, etc.
Our proposed method, although straightforward, has demonstrated exceptional performance on large-scale datasets such as SportsMOT \cite{SportsMOT} and DanceTrack \cite{dancetrack}, where varying motions and uniform appearances are present. 
\section{Related work}
\textbf{Multi-object Tracking.}
Early research on multi-object tracking mainly relied on optimization algorithms \cite{roshan2012gmcp,wen2014,lan} to solve the data association problem. However, with the advent of deep learning, tracking-by-detection using stronger detectors has become the dominant paradigm in multi-object tracking. In recent years, some approaches have fused detection and tracking tasks into a single network, benefiting from the success of multi-task learning \cite{zhang2021survey} in neural networks.
For instance, Tracktor \cite{Tracktor} predicts an object's position in the next frame by utilizing Faster RCNN's \cite{fasterrcnn} regression head, but it may fail at low frame rates. JDE \cite{jde} extends YOLOv3 \cite{yolov3} with a ReID branch to obtain object embedding for data association. To address the problem of detection and ReID tasks competing with each other in the JDE paradigm, Zhang \textit{et al.} \cite{fairmot} designed FairMOT based on an anchor-free object detector, Centernet \cite{centernet}, achieving better tracking results.
Additionally, ByteTrack \cite{bytetrack} demonstrated that the performance bottleneck of trackers on mainstream multi-object tracking datasets, MOT \cite{mot16,mot20}, is not in the association part but in the detection part. Thus, a powerful detector coupled with a simple hierarchical association strategy can achieve good tracking results.

\textbf{Motion model.}
Motion estimation is crucial for object trackers, and in the early days, many classical multi-object tracking algorithms, such as SORT \cite{sort}, DeepSORT \cite{deepsort}, and MOTDT \cite{motdt}, use Kalman filters (KF) \cite{kalman} to predict the inter-frame position offset of each object. However, the KF model is limited by its constant velocity assumption and performs poorly in complex situations and nonlinear motion patterns.

To cope with these challenges, researchers have proposed various data-driven models. For example, Zhang \etal \cite{FFT} used optical flow to obtain pixel-level motion information of objects, while CenterTrack \cite{centertrack} added an offset branch to an object detector to predict the motion information of the object center. Milan \etal \cite{milan2017} proposed an online tracker based on recurrent neural networks (RNNs) for multi-object tracking, and subsequent studies \cite{rnn1, rnn2, rnn3} applied RNNs to either fuse visual and motion information or calculate affinity scores for subsequent data association. ArTIST \cite{artist} and DEFT \cite{deft} used RNNs to predict the inter-frame motion of the target directly. TMOH \cite{TMOH} extended Tracktor \cite{Tracktor} to use a simple linear model to predict the location of lost tracks, replacing visual cues for trajectory retracking.

Recent studies have also improved the KF to handle nonlinear motion better. For example, MAT \cite{mat} proposed an IML module that considers both camera and pedestrian motion information, achieving better performance than the vanilla KF. Cao \etal \cite{ocsort} recognized the limitations of the original KF in dealing with occlusion and nonlinear motion and proposed corresponding improvements. They suggested that the KF should trust the recent observations more.

Despite these improvements, most of these studies are still based on the classical KF and its constant velocity model assumption. QuoVadis \cite{QuoVadis} improves the robustness of existing advanced trackers against long-term occlusions based on trajectory prediction in a bird's eye view (BEV) scenario representation, consisting of several complex sub-modules that overcome the limitations of traditional methods.

\textbf{Transformer-based methods.}
The recent success of transformer models in computer vision\cite{detr,defdetr,WANG2024106110,CAI2024548}, particularly in the field of object detection, has led to the emergence of numerous transformer-based approaches. These include TransTrack \cite{transtrack}, TrackFormer \cite{trackformer}, TransCenter \cite{transcenter}, and MOTR \cite{motr}, which are contemporaneous online trackers based on DETR \cite{detr} and its variants. TrackFormer employs track queries to maintain object identities and utilizes heuristics to suppress duplicate tracks, as in Tracktor \cite{Tracktor}. TransTrack directly employs previous object features as track queries to acquire tracking boxes and associates detection boxes based on IoU-matching. TransCenter obtains object center representation through transformer and performs tracking through CenterTrack's object association \cite{centertrack}. Additionally, MOTR performs object tracking in an end-to-end manner by iteratively updating the track query, without requiring post-processing, making it a concise method. Furthermore, GTR \cite{gtr} is an offline transformer-based tracker that employs queries to divide detected boxes into trajectories all at once, instead of generating tracking boxes. It should be noted that the training of all these models necessitates a high volume of training samples, expensive computational resources, and lengthy training times. In contrast, our methodology exclusively utilizes the Transformer to leverage object trajectory information. Furthermore, our proposed method requires only trajectory data as input and is distinguished by its rapid training process.

\begin{figure*}[t]
  \centering
   \includegraphics[width=1.0\linewidth]{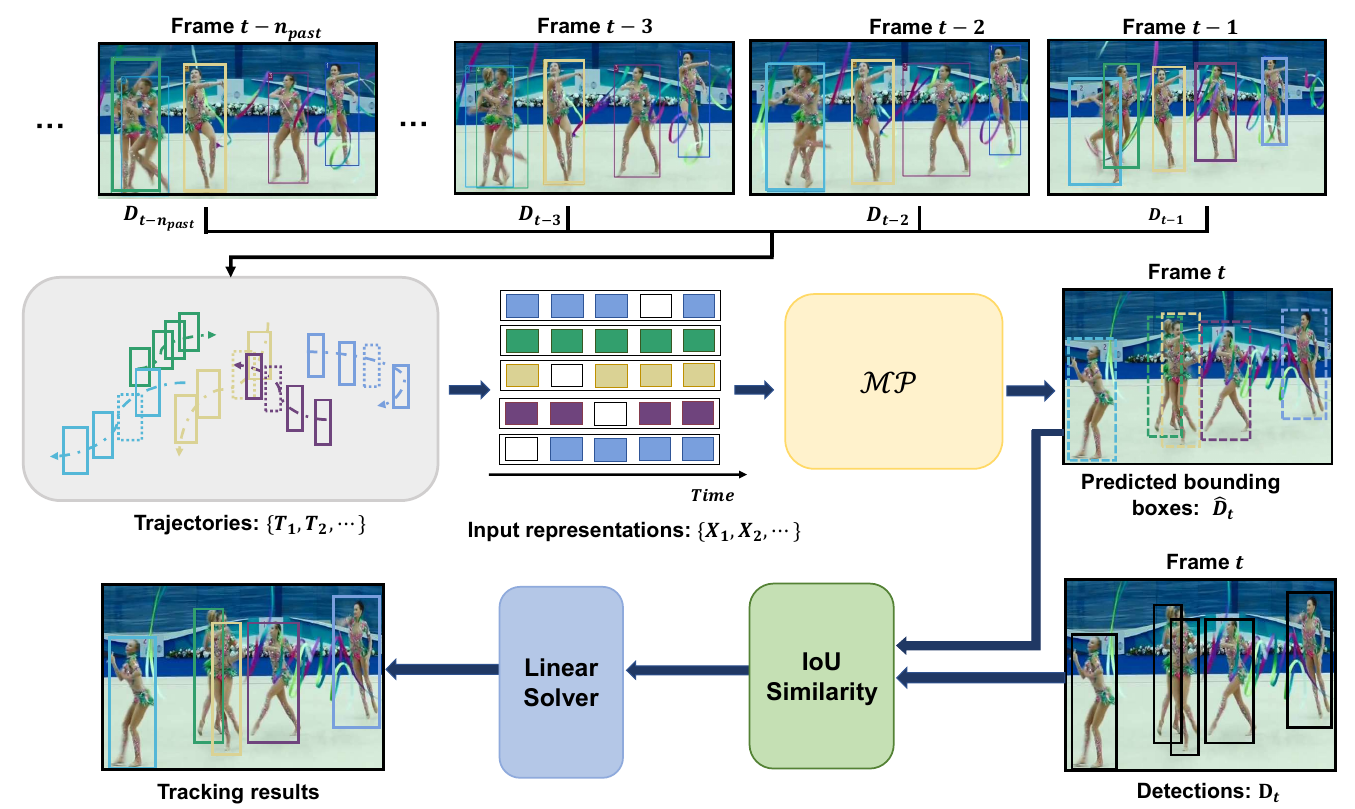}
  \caption{An overview of the proposed method. The proposed motion predictor $\mathcal{MP}$ considers at most $n_{past}$ of the historical observations of its trajectory when predicting the object position. With predicted bounding boxes $\hat{\bold{D}}_t$, data association can be achieved by the linear solver, Hungarian algorithm, based solely on their spatial similarity to the current frame detection results $\bold{D}_t$. Blank boxes represent missing observations and dashed boxes represent predicted bounding boxes. Different colors represent different objects.}
  \label{fig:overview}
\end{figure*}

\section{Method}
\label{sec:method}

The multi-object tracking task involves identifying the spatial and temporal locations of objects, i.e., their trajectories, in a given video sequence. The Transformer model, known for its ability to capture long-term dependencies, has proven highly effective in processing sequence data. Building on this success, we present a motion predictor that utilizes information about an object's past trajectory information to predict its position in the next frame directly, as illustrated in \Cref{fig:overview}. Our data association approach relies solely on spatial similarity between the detection results of the current frame and the predicted bounding box of the object.
To achieve this, we initially introduce a simple base model, outlined in Section~\ref{sec:predictor}, based on the vanilla Transformer encoder to capture the temporal dynamics of individual objects. Following that, we introduce the Dynamic MLP, an MLP-like architecture designed to explore channel-wise interactions.
We further integrate it with the self-attention module to learn granularity information across token and channel levels in Section \ref{sec:mif}. Finally, we study various augmentations to construct more complex motion patterns to improve the understanding of motion dynamics.
Overall, our approach leverages the strengths of the Transformer model and builds on it to develop a simple and effective motion predictor for multi-object tracking. 









\subsection{Notations}
\label{sec:Notations}
The trajectory of an object consists of an ordered set of bounding boxes $ \mathcal{T} = \{\textbf{b}_{t_1}, \textbf{b}_{t_2,}, \dots \}$, where a bounding box is defined as $ \textbf{b}_t = \{ x, y, w, h\} $, and $t$ stands for the timestamp. At some moments, observations may be missing due to occlusion or detector failure, so the trajectory is not always continuous. $ \textbf{D}_t $ is the set of detections of the $t$-th frame provided by an off-the-shelf detector.  The past trajectory representation of the object can be denoted as a sequence 
$\textbf{X} = ( \dots, \textbf{x}_{t-2}, \textbf{x}_{t-1}) \in \mathbb{R}^{n \times 9}$.
The representation of an object at moment $t-1$ is denoted as follows:
\begin{equation}
  \textbf{x}_{t-1} = (c_x, c_y, w, h, a, \delta _{c_x}, \delta _{c_y}, \delta_w, \delta_h),
  \label{eq:x_def}
\end{equation}
where $(c_x, c_y)$ represents the center coordinate of the object in the image plane,  $w$, $h$ and $a$ stand for width, height and aspect ratio of its bounding box respectively. $\delta _{c_x}$, $\delta _{c_y}$, $\delta_w$, and $\delta_h$ denote the variation in the center position, width, and height relative to the previous observation.

For each individual object, the motion predictor, denoted as $\mathcal{MP}$, leverages up to $n_{past}$ of its historical observations, represented by $\textbf{X}_{t-n_{past}:{t-1}}$, to predict the positional offset, denoted as $\boldsymbol{\hat{O}}_t$, in the subsequent frame.
The objective of object motion prediction is to forecast the relative spatial displacement of the object bounding box based on its historical trajectory information:
\begin{equation}
   \begin{split}
   \textbf{X}_{t-n_{past}:{t-1}} &= \text{Concat}(\textbf{x}_{t-n_{past}}, \dots, \textbf{x}_{t-1}), \\
   \boldsymbol{\hat{O}}_t &=  \mathcal{MP}(\textbf{X}_{t-n_{past}:{t-1}}),
   \end{split}
\end{equation}

Before being fed into the encoder layer for further processing, $\textbf{X}$ is embedded onto a higher dimensional space by a linear projection, i.e., $ \Bar{\textbf{X}} = \textbf{W}_x\textbf{X} \in \mathbb{R}^{n \times d_m}$.
In order to make the input token sequence $ \textbf{E} \in \mathbb{R}^{n \times d_m}$ contain relative position information, we inject sinusoidal position encoding information to the input embeddings as in \cite{transformer}.

\subsection{Vanilla Transformer-based Motion Predictor}  
\label{sec:predictor}



The future motion of an object is significantly influenced by its past dynamic information. An intuitive approach utilises a vanilla Transformer encoder to capture the historical context of individual objects flexibly and efficiently. This encoder can model the long-term dependencies within the object's trajectory history, where the primary component is the multi-head self-attention (MHSA) mechanism. By employing MHSA, the encoder can efficiently attend to various elements of the trajectory sequence and identify the most informative features that contribute to predicting the future motion of the object.

We compute the attention for every single object separately. The input sequence of tokens $\textbf{E}$ is linearly transformed to $Q$, $K$, $V$, which are the query, key, and value, respectively. For multi-head self-attention, a set of single-head attention jointly attends to information from different representation subspaces. The attention of a single head is calculated as by:

\begin{equation}
  Attention(Q,K,V) = softmax(\frac{QK^T}{\sqrt{d_k}})V,
  \label{eq:attention}
\end{equation}
where $d_k$ is the dimensionality of the corresponding hidden representation as a scaling factor. 
The multi-head attention variant with $h$ heads, whose outputs are denoted as $head_0, head_1,\dots, head_i, \dots, head_h$,
\begin{equation}
    head_i = Attention_i(Q_i,K_i,V_i),
    \label{eq:mhsa}
\end{equation}
where $Q_i$, $K_i$, $V_i$ are the $i_{th}$ part of $Q$, $K$ and $V$. The outputs of $h$ heads are cascaded and linearly transformed to obtain the final output.

\label{sec:multi}
Multi-object tracking is challenging because the interactions between objects, i.e., the effect of one object's behaviour on other objects, is a sophisticated process. Like some of the previous work\cite{artist,relationtrack,braso2020learning}, we also tried to model ``interactions'' between objects. 
We follow the Agentformer\cite{yuan2021agentformer} and use an agent-aware attention mechanism to model multi-object motion in both temporal and social dimensions using sequential representations. We denote this variant as \textit{mult}.


\begin{figure*}[!t]
    \centering
     \centering
    \subfloat[]{\includegraphics[width=0.4\linewidth]{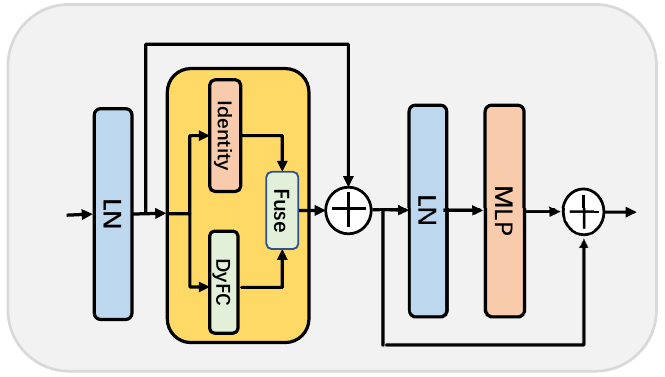}}
    \label{fig:dymlp_a}
   \hfill
    \subfloat[]{\includegraphics[width=0.5\linewidth]{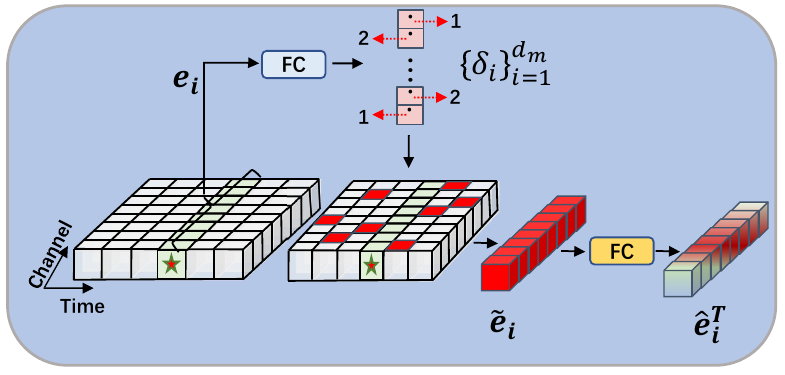}}
    \label{fig:dymlp_b}
  \caption{The network structure of dynamic MLP is shown in (a), and the dynamic FC operation process is shown in (b).}
  \label{fig:dymlp} 
\end{figure*}

\subsection{Dual-granularity Information Fusion}
\label{sec:mif}


The utilization of token-level message passing across all channels alone lacks the capacity to adequately convey divergent semantics. This limitation restricts their optimal exploitation, consequently leading to sub-optimal performance\cite{bau2020understanding,wu2021stylespace, zhang2023crossformer,chen2023tsmixer}. 
Therefore, there is a need to employ a more fine-grained message passing mechanism to cater to varying semantics in an adaptive manner. 
Such an approach can lead to more sufficient motion modeling by allowing for finer granularity in the communication of information cross channels.

\paragraph{Dynamic MLP} We explore a new module named Dynamic MLP (DyMLP) in parallel with self-attention in vanilla transformer encoder layer to further capture complicated motion patterns of objects. It aggregates multiple positions distributed in different temporal channels. As shown in \cref{fig:dymlp}, the core of DyMLP is the channel fusion layer (CFL), which consists of a dynamic fully-connected layer (DyFC) and an identity layer. Given the input token sequence $ \textbf{E} \in \mathbb{R}^{n \times d_m}$, for each token $e_i \in \mathbb{R}^{d_m}$, we first use a FC layer to predict $d_m$ offsets: $\Delta = \{ \delta_i \}_{i=1}^{d_m}$. Since there is no restriction on the generation of offsets, DyFC can aggregate temporal global channel information, as shown in \cref{fig:dymlp}. The basic DyFC operator can be formulated as below:

\begin{equation}
  \begin{split}
  \hat{e}^T_i = {\rm DyFC}(e_i) = \tilde{e}_i \cdot \textbf{W} + \textbf{b},\\
    \tilde{e}_i = [\textbf{E}_{[i+ \delta_1, 1]}, \textbf{E}_{[i+ \delta_2, 2 ]},\cdots, \textbf{E}_{[i+ \delta_{d_m}, d_m ]}],
  \end{split}
  \label{eq:dyfc}
\end{equation}
where $\textbf{W} \in \mathbb{R}^{d_m \times d_m} $ and $\textbf{b} \in \mathbb{R}^{d_m}$ are learnable parameters.
Besides, we preserve the original token information using identity layer and the output is $\hat{e}^I_i$. 
CFL outputs the fusion result as a weighted sum of $\hat{e}^T$ and $\hat{e}^I$, which can be formulated as:
\begin{equation}
    \hat{e} = \omega^T \odot \hat{e}^T + \omega^I \odot \hat{e}^I,
    \label{eq:fuse}
\end{equation}
where $\odot$ is the Hadamard product, $\omega^{\{T,I\}} \in \mathbb{R}^{d_m}$ are calculated from the following equation:

\begin{equation}
    [\omega^I, \omega^T] = softmax([W^I \cdot \dot{x}, W^T \cdot \dot{x}]),
\end{equation}
where $\dot{x} \in \mathbb{R}^{d_m}$ is the average summation of $\hat{e}^T$ and $\hat{e}^I$, $W^{\{I,T\}} \in \mathbb{R}^{d_m \times d_m}$ are learnable parameters and $softmax(\cdot)$ is the channel-wise normalization operation.


Hence, each encoder layer contains two sub-layers.  We use residual connections \cite{he2016deep} in both sub-layers and then perform layer normalization (LN) \cite{ba2016layer}.  Mathematically, the whole process in the encoder layer can be described as follows:
\begin{equation}
    \begin{split}
     \text{DIF}(E^{l-1}) &= \text{MHSA}(E^{l-1}) + \text{DyMLP}(E^{l-1}) \\
    \hat{E}^l  &= \text{LN} (\text{DIF}(E^{l-1})) + E^{l-1} \\
    E^l  &= \text{LN} (\text{FFN}(\hat{E}^l)) + \hat{E}^l,
    \end{split}
\end{equation}
where $l$ denotes the $l$-th layer and FFN for a feed forward network. 




\begin{figure}[!t]
    \centering
    \includegraphics[width=0.7\linewidth]{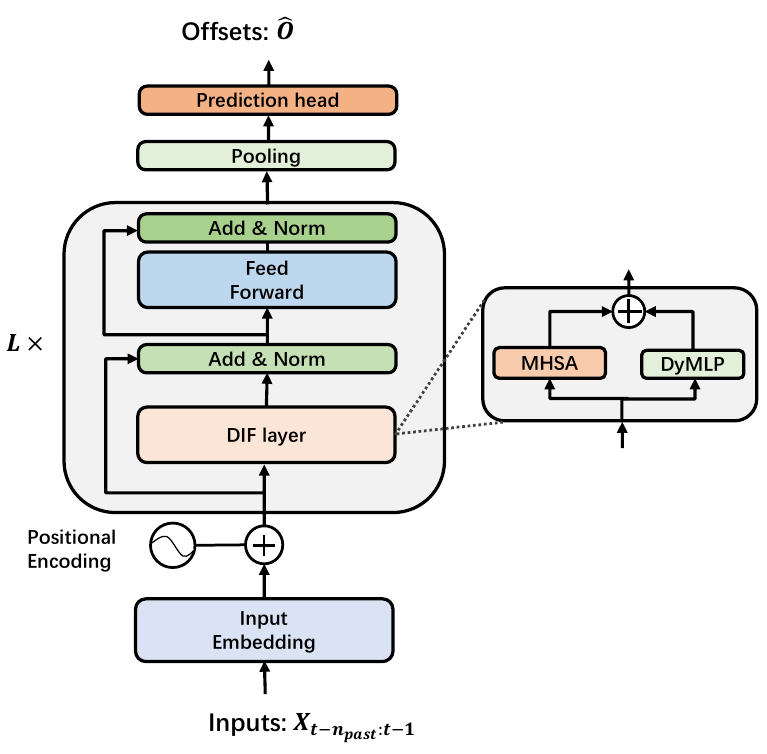}
  \caption{The architecture of the proposed motion predictor. }
  \label{fig:motion_predictor} 
\end{figure}

\paragraph{Dual-granularity Information Fusion Layer}

With the introduction of the proposed Dynamic MLP, we integrate it with a self-attention module in Transformer \cite{transformer}. This integration leverages information from both token-level granularity and channel-level granularity. We refer to this newly combined module as the Dual-granularity Information Fusion Layer (DualIF). Illustrated in \cref{fig:motion_predictor}, the backbone of the motion predictor $\mathcal{MP}$ comprises $L$ sequentially connected encoding layers. Each encoding layer is composed of two sub-modules: a multi-head self-attention module and a dynamic MLP module. The former incorporates token-level information from other tokens into the query based on computed pairwise attention weights, while the latter adaptively performs feature fusion of context distributed at the channel level.

\subsection{Training and inference}
\label{sec:training_inf}

\paragraph{Data augmentation creates complex motion patterns}
Like in other deep learning tasks, such as Centernet \cite{centernet}, Centertrack \cite{centertrack}, and YOLOX \cite{yolox}, the incorporation of data augmentation is crucial in enhancing the performance of the model. The limited availability of training samples within existing datasets, coupled with the variability of motion across diverse datasets, necessitates the exploration of novel augmentation strategies to augment the training samples and advance the proficiency of our predictor in modeling motion dynamics. In this study, we examine three distinct augmentation strategies aimed at enriching the training samples. These strategies correspond to various cases that must be handled by the motion model, namely object position jitter and motion mutations, detection noise, and newly initialized shorter trajectories.


\begin{itemize}
  \item Random drop. In generating the target trajectory representation \textbf{X}, we randomly ignore the observation \textbf{b} with probability $\textit{p}_i$. This strategy allows to simulate fast motion and low frame rate scenes.
  \item Spatial jitter. To enhance the robustness of the system against detection noise, it is a widely adopted practice to incorporate spatial jitter into the bounding boxes. This technique involves introducing small variations in the position and size of the bounding boxes during training. By doing so, the model becomes more resilient to variations in object localization due to detection noise. Furthermore, in a temporal sense, this approach also serves to augment the training dataset with a broader range of motion patterns.
  \item Random length. Due to the short new birth trajectory, there is little available temporal information. By randomly varying the length of the training sequence, the model is exposed to a diverse range of motion patterns, which enables it to learn to generalize to different temporal contexts. To mimic this, we train the motion model using a sequence of observations of arbitrary length (in range [2, $\textit{n}_{past}$]) reserved for each object.
\end{itemize}

\paragraph{Training loss} 
We adopt the smooth loss L1 \cite{fastrcnn} to supervise the training process. Formally, given the predicted offsets, $ \boldsymbol{\hat{O}} = \{ \delta _{c_x}, \delta _{c_y}, \delta_w, \delta_h \}$, and corresponding ground truth $ \boldsymbol{O} $, the loss is obtained by
\begin{equation}
    L(\hat{O}, O) = \sum_{i\in \{c_x, c_y, w, h\}}{ \text{smooth}_{L_1}(\hat{\delta}_i - \delta_i), }
\end{equation}
in which
\begin{equation}
    \text{smooth}_{L_1}(x) = 
    \left\{ 
    \begin{aligned}
        &0.5x^2                  & \quad \text{if} \left| x \right| < 1 \\
        &\left| x \right| - 0.5  & \quad \text{otherwise}. \\
    \end{aligned}
    \right.
\end{equation}

\paragraph{Inference} 
At first, we decode the predicted offset $\hat{\textbf{O}}_t$ as the trajectory bounding boxes $\hat{\textbf{D}}_t$ in the current frame. 
As with some classic online trackers\cite{sort,deepsort,motdt,bytetrack}, we exploit a simple association algorithm. Detections are assigned to tracklets based on Intersection-over-Union(IoU) similarity between $\hat{\textbf{D}}_t$ and $\textbf{D}_t$ using the Hungarian algorithm\cite{hungarian}. Unassigned detections are initialized as new trajectories. If no detection is assigned to a trajectory, the trajectory is marked as lost and if the time lost is greater than a given threshold, the target is considered out of view and removed from the trajectory set. Lost trajectories may also be retracked in the assignment step.

Given the detections from an object detector, our tracker associates identities over video sequences in an online manner, exploiting only motion cues. The overall tracking pipeline is shown in Algorithm \ref{alg:alg_track}. For brevity, trajectory rebirth is not shown.

\begin{algorithm}[h]
 \small
 \caption{Pseudo-code of MotionTrack.}
 \label{alg:alg_track}
 \KwIn{ Detections: $\textit{D}= \{\textbf{b}_t^i| 1 \leq t \leq M, 1 \leq i \leq N_t\}$, Motion Predictor: $\mathcal{MP}$,  threshold for retaining a lost track $t_{max}$.}
 \KwOut{Tracks $\textit{T}$ of the video}
 \BlankLine
 Initialization: $\textit{T} \leftarrow \emptyset $ and $\mathcal{MP}$; \\
 \For{$t \leftarrow 1:M$}
    {
    $\textbf{D}_t \leftarrow [\textbf{b}_t^1, \cdots, \textbf{b}_t^{N_t}]$ \tcp{Detections of current frame.}
    $\hat{\textbf{D}}_t \leftarrow [\hat{\textbf{b}}_t^1, \cdots, \hat{\textbf{b}}_t^{|\textit{T}|}]$  from $\textit{T}$ \tcp{Predicted bounding boxes}
    $\mathbf{C}_t \leftarrow C_{\rm IoU} (\hat{\textbf{D}}_t, \textbf{D}_t)$ \tcp{Cost matrix based on IoU similarity}
    \tcc{Assign detections to tracks using Hungarian algorithm}
    $\mathcal{M},\mathcal{T}_u,\mathcal{D}_u \gets \rm assignment(\mathbf{C}_t)$  \\
    $\textit{T} \leftarrow \{T_i(\textbf{b}_t^j), \forall (i,j) \in \textit{M}\}$ \tcp{Update the matched tracks}
    $\textit{T} \leftarrow \{T_i.age += 1, \forall (i) \in \textit{T}_u\}$ \\
    $\textit{T} \leftarrow \{T_i(D_j), \forall {j} \in \textit{D}_u, i=|\textit{T}| + 1 \}$ \\
    \textbf{Kill lost tracks with age $\geq t_{max}$} \\
    
    \For{$T$ in $\textit{T}$}
        {$\mathcal{MP}$($T$) \tcp{Predict motion of tracks}
        }  
 }  
\end{algorithm}   

\section{Experiments}
\label{sec:experiments}

\subsection{Datasets and Metrics}
\myPara{Datasets.} 
In order to conduct a comprehensive evaluation of the proposed algorithm, we performed experiments on two datasets characterized by complex motion patterns, namely DanceTrack\cite{dancetrack} and SportsMOT \cite{SportsMOT}.  The SportsMOT dataset offers video clips of three different sports categories, i.e., basketball, football, and volleyball. These clips are collected from various sources such as the Olympic Games, NCAA Championship, and NBA on YouTube, and cover a wide range of complex sports scenes captured from different perspectives. The dataset contains 45 video clips in both the training and validation sets. Due to the similar appearance of athletes and the complex motion scenarios, SportsMOT requires a high level of robustness in tracking algorithms.
Dancetrack, on the other hand, is a recently proposed dataset that offers more training and evaluation videos. Objects in this dataset possess highly similar appearance, are severely occluded from each other, and exhibit nonlinear motion at a high degree, making it challenging for existing advanced approaches based on appearance and motion information. Therefore, we aim to propose a better motion model that can improve the tracking algorithm's ability to cope with frequent crossings and nonlinear motions. SportsMOT and Dancetrack datasets provide ideal benchmarks for evaluating the performance of tracking algorithms.

\myPara{Evaluation Metrics.} To evaluate our algorithm, we adapt the Higher Order Metric (HOTA, AssA, DetA) \cite{hota}, IDF1\cite{idf1} and the CLEAR metrics (MOTA, FP, FN, IDs, \etal) \cite{clear} 
to assess different aspects of the tracking algorithm. MOTA is calculated from FN, FP, IDs and is susceptible to the influence of detection results. IDF1 focuses on measuring association performance. HOTA is designed to fairly combine evaulation of detection and association, and therefore, we use it as the primary metric.

\subsection{Implementation Details}
Our primary focus is on developing a motion model for tracking objects. For this purpose, we use the publicly available YOLOX detector weights provided by ByteTrack \cite{bytetrack}, DanceTrack \cite{dancetrack} and SportsMOT\cite{SportsMOT} separately to detect objects in the MOT, DanceTrack and SportsMOT datasets. The motion predictor includes $L=6$ encode layers, and the input token dimension $d_m$ is set to 512. The multi-head self-attention uses 8 heads, and the drop probability for data augmentation during training is set to $\textit{p}_i = 0.1$. We set the maximum historical observation window value $\textit{n}_{past}$ to 10, and the batch size to 64. We use the Adam optimizer \cite{adam} with $\beta_1 = 0.9$, $\beta_2 = 0.98$ and $\epsilon = 10^{-8}$. The learning rate is adjusted dynamically during training following the approach proposed in \cite{transformer}.

\begin{table}[!t]
\centering
\setlength{\tabcolsep}{0.8mm}
\caption{Comparison to the state-of-the-arts on DanceTrack test set. The best results are shown in \textbf{bold}.↑: higher better.}
\begin{tabular}{|l|c|c|ccccc|}
\hline
\textbf{Tracker}               & \textit{Motion}     & \textit{Appear.}    & HOTA↑         & DetA↑         & AssA↑         & MOTA↑         & IDF1↑         \\ \hline
DeepSORT \cite{deepsort}       & \checkmark & \checkmark & 45.6          & 71.0          & 29.7          & 87.8          & 47.9          \\
MOTR \cite{motr}               & \checkmark & \checkmark & 54.2          & 73.5          & 40.2          & 79.7          & 51.5          \\
FairMOT \cite{fairmot}         & \checkmark & \checkmark & 39.7          & 66.7          & 23.8          & 82.2          & 40.8          \\ \hline
QDTrack \cite{quasi}           &            & \checkmark & 45.7          & 72.1          & 29.2          & 83.0          & 44.8          \\
TransTrk \cite{transtrack}     &            & \checkmark & 45.5          & 75.9          & 27.5          & 88.4          & 45.2          \\
TraDes \cite{wu2021track}      &            & \checkmark & 43.3          & 74.5          & 25.4          & 86.2          & 41.2          \\ \hline
CenterTrack \cite{centertrack} & \checkmark &            & 41.8          & 78.1          & 22.6          & 86.8          & 35.7          \\
SORT \cite{sort}               & \checkmark &            & 47.9          & 72.0          & 31.2          & \textbf{91.8} & 50.8          \\
ByteTrack \cite{bytetrack}     & \checkmark &  & 47.3          & 71.6          & 31.4          & 89.5          & 52.5          \\
OC\_SORT \cite{ocsort}         & \checkmark &            & 55.1          & 80.3          & 38.0          & 89.4          & 54.2          \\
Ours                           & \checkmark &            & \textbf{58.2} & \textbf{81.4} & \textbf{41.7} & 91.3          & \textbf{58.6} \\ \hline
\end{tabular}
\label{table:dance_test}
\end{table}

\begin{table}[!t]
\centering
\setlength{\tabcolsep}{0.8mm}
\caption{Comparison to the state-of-the-arts on SportsMOT test set. The best results are shown in \textbf{bold}.↑: higher better.}
\begin{tabular}{|l|c|c|ccccc|}
\hline
Tracker                         & \textit{Motion}     & \textit{Appear.}    & HOTA$\uparrow$ & DetA$\uparrow$ & AssA$\uparrow$ & MOTA$\uparrow$ & IDF1$\uparrow$ \\ \hline
FairMOT\cite{fairmot}           & \checkmark & \checkmark & 49.3           & 70.2           & 34.7           & 86.4           & 53.5           \\
MixSort-Byte \cite{SportsMOT} & \checkmark & \checkmark & 65.7           & 78.8           & 54.8           & 96.2           & 74.1           \\
MixSort-OC \cite{SportsMOT}   & \checkmark & \checkmark & 74.1           & 88.5           & 62.0           & 96.5           & 74.4           \\ \hline
QDTrack\cite{quasi}             &            & \checkmark & 60.4           & 77.5           & 47.2           & 90.1           & 62.3           \\
TransTrack\cite{transtrack}     &            & \checkmark & 68.9           & 82.7           & 57.5           & 92.6           & 71.5           \\
GTR\cite{gtr}                   &            & \checkmark & 54.5           & 64.8           & 45.9           & 67.9           & 55.8           \\
CenterTrack\cite{centertrack}   & \checkmark &            & 62.7           & 82.1           & 48.0           & 90.8           & 60.0           \\
ByteTrack\cite{bytetrack}       & \checkmark &            & 62.8           & 77.1           & 51.2           & 94.1           & 69.8           \\
OC-SORT\cite{ocsort}            & \checkmark &            & 71.9           & 86.4           & 59.8           & 94.5           & 72.2           \\
Ours                            & \checkmark &            & \textbf{74.0}           & \textbf{88.8}           & \textbf{61.7}          & \textbf{96.6}           & \textbf{74.0}          \\ \hline
\end{tabular}
\label{table:sports_test}
\end{table}

\subsection{Benchmark Evaluation}
We assessed the efficacy of the proposed method by testing it on two datasets comprising various motion patterns. 

\myPara{DanceTrack.}
Table \ref{table:dance_test} presents a comparison between our proposed method and the current state-of-the-art techniques on the Dancetrack test set. 
The data presented in the table illustrates that our method, rely solely on motion cues, exhibits superior performance in comparison to a spectrum of algorithms that integrate both motion and appearance cues, or either of these tracking cues in isolation.
Specifically, our method achieves a higher HOTA score, which is 3.1 percentage points better than the OC\_SORT\cite{ocsort} method that enhances the Kalman filter. Furthermore, in the evaluation metrics that concentrate on measuring tracking performance, namely AssA and IDF1, our approach leads by 3.7\% and 4.4\%, respectively. These findings suggest that our proposed data-driven motion model outperforms SORT-like techniques \cite{sort,deepsort,bytetrack} based on the standard Kalman filter \cite{kalman}. Additionally, our method surpasses the advanced fully end-to-end MOTR\cite{motr} method. These results provide evidence that our proposed motion model can effectively model the temporal motion of objects and achieve superior performance in dealing with complex motion scenes.

\myPara{SportsMOT.} 
SportsMOT is a newly proposed dataset, which is mainly characterized by the variety of object motion patterns in the dataset, but at the same time their appearance is similar but distinguishable. As shown in \ref{table:sports_test}, our approach is surpassed by all methodologies which only relying on either motion or appearance cues for tracking. More specifically, our method outperforms OC\_SORT \cite{ocsort} by 2.1 percentage points on the HOTA metric and by 1.9 percentage points on the AssA metric. Furthermore, our methodology demonstrates comparable performance to state-of-the-art hybrid tracking algorithms that concurrently leverage both appearance and motion information. The excellent performance obtained on the SportsMOT dataset serves to underscore the proficiency of our algorithm in accurately modeling dynamic and variable-speed motion, thereby furhter enabling precise tracking of athletes across a diverse range of sport scenarios.

\begin{figure*}[!h]
  \centering
  \includegraphics[width=1.0\linewidth]{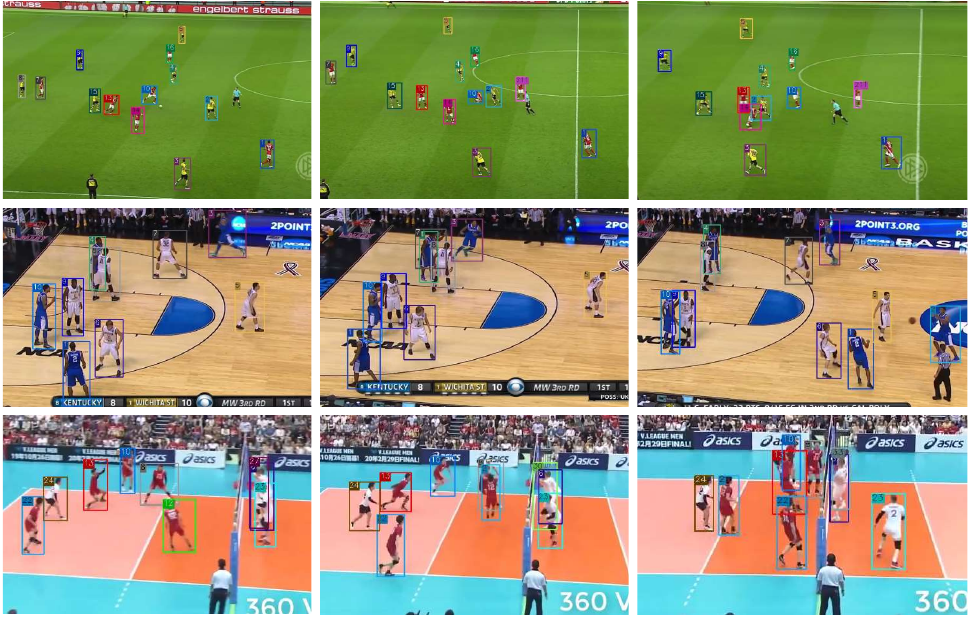}
   \caption{Qualitative results of our method on SportsMOT. Different colored bounding boxes indicate different identity. Best viewed in color and zoom in.}
  \label{fig:sportsval} 
\end{figure*}

\begin{figure*}[!t]
    \centering
    \subfloat[OC\_SORT: dancetrack0004.]{\includegraphics[width=0.48\linewidth, height=4.3cm]{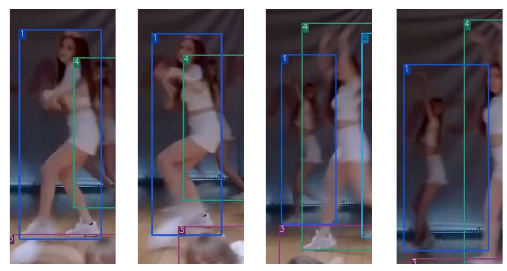}}
    \label{fig:short-aa}
     \hfill
    \subfloat[MotionTrack: dancetrack0004.]{\includegraphics[width=0.48\linewidth,height=4.3cm]{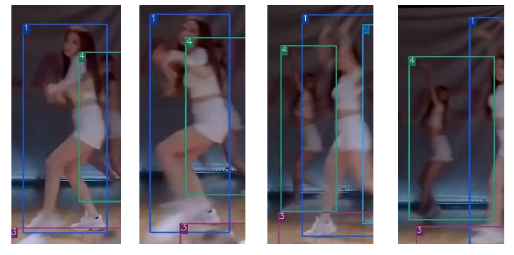}}
    \label{fig:short-bb}
     \subfloat[OC\_SORT: dancetrack0097.]{\includegraphics[width=0.48\linewidth, height=4.3cm]{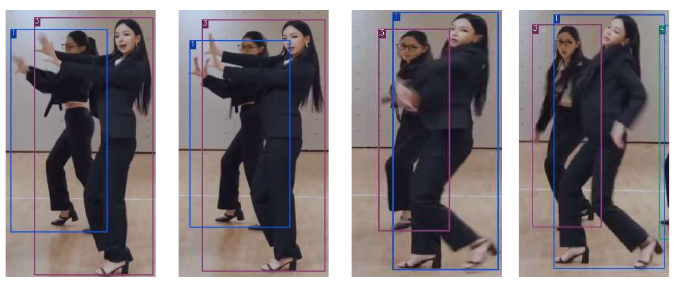}}
    \label{fig:short-aaa}
  \hfill
    \subfloat[MotionTrack: dancetrack0097.]{\includegraphics[width=0.48\linewidth, height=4.3cm]{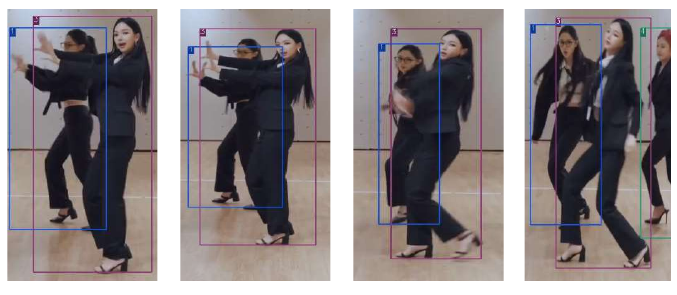}}
    \label{fig:short-bbb}
   \hfill
    \subfloat[OC\_SORT: dancetrack0094.]{\includegraphics[width=0.48\linewidth, height=4.3cm]{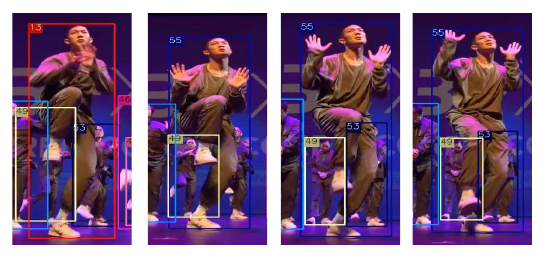}}
    \label{fig:short-ccc}
     \hfill
    \subfloat[MotionTrack: dancetrack0097.]{\includegraphics[width=0.48\linewidth, height=4.3cm]{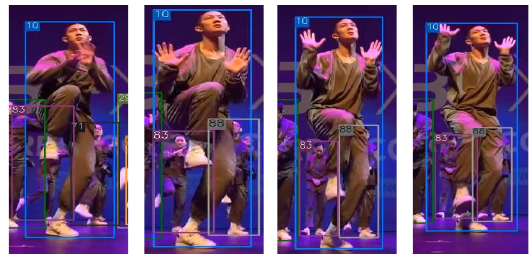}}
  \caption{Qualitative comparison between MotionTrack and OC\_SORT\cite{ocsort} on DanceTrack validation. Each pair of rows shows the results comparison for one sequence. The color of the bounding boxes represents the identity of the tracks. To be exactly, the ID switch in (a) occurs between frame \#45 $\xrightarrow{}$ \#54;
  (c) \#310 $\xrightarrow{}$ \#333; 
  (e) \#432 $\xrightarrow{}$ \#441.} 
  \label{fig:comparision_1} 
\end{figure*}

\myPara{Qualitative Results.}
To get a more intuitive picture of MotionTrack's superiority over OC\_SORT, we provide more visualization for the comparison. We show some qualitative results of our method on SportsMOT in \cref{fig:sportsval}. In addition, in \ref{fig:comparision_1}, we show additional samples where OC\_SORT suffers from ID switch caused by nonlinear motion or  occlusion but our method successfully copes.

\subsection{Ablation Study}

We conduct ablation studies on the validation set of DanceTrack \cite{dancetrack} to investigate the impact of different training data, model components, data augmentation methods, and some hyper-parameters on the proposed method.



\begin{figure*}[!t]
  \centering
   \includegraphics[width=0.8\linewidth]{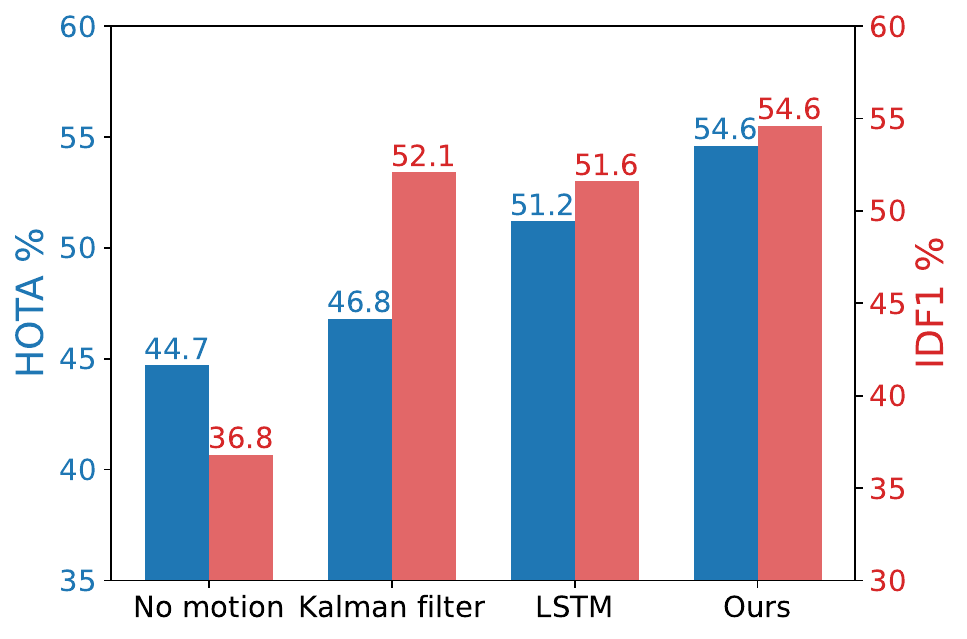}
  \caption{Comparing different motion models.}
  \label{fig:motion}
\end{figure*}

\myPara{Impact of motion modelling.} 
To validate the efficacy of the proposed motion predictor, we conducted a comparative analysis with several existing motion models. As illustrated in \Cref{fig:motion}, the outcomes demonstrate that incorporating motion information yields significantly better performance than those not using motion information (i.e. only naive IoU association). 
Furthermore, the HOTA metrics reveal that our approach outperforms Kalman filter \cite{kalman}, which relies on linear motion hypothesis, by 7.8\%, and LSTM \cite{deft}, which employs a recurrent neural network, by 3.4\%, respectively. This notable improvement in performance is attributed to the dual-granularity information incorporated in our motion model, which provides a global temporal perspective and thereby improves motion learning.

\begin{table}[!t]
\centering
\caption{Ablation study on different components. ``MHSA'' stands for multi-head self-attention, ``DyMLP'' for dynamic MLP.}
\setlength{\tabcolsep}{2.8mm}
\begin{tabular}{|l|ccccc|}
\hline
                   & HOTA↑ & DetA↑ & AssA↑ & MOTA↑ & IDF1↑  \\
\hline   
\textit{w/o} MHSA      & 53.6  & 78.0  & 37.0  & \textbf{89.2}  & 54.1 \\
\textit{w/o} DyMLP     & 53.0  & \textbf{78.6}  & 35.8  & 89.1  & 53.2 \\
Ours                  & \textbf{54.6}  & \textbf{78.6}  & \textbf{38.1}  & \textbf{89.2}  & \textbf{54.6}  \\
\hline
\end{tabular}
\label{table:components}
\end{table}

\begin{table}[!t]
\centering
\caption{Evaluation of the impact of inter-object interactions. \textit{Mult.} denotes the method of modeling inter-target interactions using agent-aware attention.}
\setlength{\tabcolsep}{2.8mm}
\begin{tabular}{|l|ccccc|}
\hline
          & HOTA↑ & DetA↑ & AssA↑ & MOTA↑ & IDF1↑  \\
\hline   
\textit{Mult.}     & 53.1  & \textbf{78.9}  & 35.9  &  \textbf{89.2} & 53.0  \\
Ours      & \textbf{54.6}  & 78.6  & \textbf{38.1}  & \textbf{89.2}  & \textbf{54.6}  \\
\hline
\end{tabular}
\label{table:mult}
\end{table}

\myPara{Impact of model components}. 
To further investigate the impact of the key components of our motion predictor, we conducted an ablation study by removing them individually and evaluating the resulting model's performance. The results of this study are presented in \Cref{table:components}.
As shown in the table, when the MHSA module and the DyMLP module are removed separately from the motion predictor, there is a decline in the HOTA metric by 1\% and 1.6\%, respectively. Moreover, the association score (AssA) also suffers a reduction of 1.1\% and 2.3\%, respectively. These findings highlight the crucial role of both token-level and channel-level feature fusion in learning motion models that capture the object's temporal dynamics effectively.

\myPara{Does the interaction between objects help?}
In the context of multi-object tracking, it is pertinent to consider object interactions due to the complexity of the task. In this part, we adopt the Encoder component of AgentFormer\cite{yuan2021agentformer} to implement an object-aware attention mechanism that jointly models the temporal and social dimensions of multi-object trajectories using a sequence representation. It is noteworthy that conventional trajectory prediction tasks typically require a complete history of the target's past trajectory to predict a fixed-length future trajectory.  To adapt this, \textit{mult} (see \cref{sec:multi}) utilizes a sequence representation of multi-object trajectories by concatenating trajectory features across time and objects.
Unlike traditional trajectory prediction tasks, the number of targets in multi-object tracking varies over time and the target trajectories may not be continuous at all time. To adapt the agent-aware attention mechanism, we employ learnable embeddings to fill in the missing objects' representation within the maximum observation window of $\textit{n}_{past}$. Consequently, the \textit{mult} can be considered as a trajectory prediction task that is consistent with AgentFormer but predicts only a single time step.
Our experimental results, as presented in \cref{table:mult}, reveal that in scenarios with multiple motion patterns created by choreographers, modeling of inter-target interaction does not enhance the performance of multi-object tracking. We must emphasize that our study only explores the potential of joint attention-based social and temporal modeling. Other methods may be effective.

\begin{table}
\centering
\caption{Ablation on different data augmentation strategies. 'D' stands for random drop, 'J' stands for random jitter and 'L' stands for random length.}
\setlength{\tabcolsep}{1.8mm}
\begin{tabular}{|c|lccccc|}
\hline
NUM                          &             & HOTA↑         & DetA↑         & AssA↑         & MOTA↑         & IDF1↑         \\ \hline
                             & No aug.     & 53.5          & 78.6          & 36.5          & 89.3          & 53.5          \\
                             & + D         & 54.3          & 78.4          & 37.8          & 89.2          & 54.2          \\
                             & + J         & 54.4          & 78.6          & 37.8          & 89.2          & 54.5          \\
 \multirow{1}{1em}{\textcircled{1}} & + D + J     & \textbf{54.6} & \textbf{78.6} & \textbf{38.1} & 89.2          & \textbf{54.6} \\
                             & + D + J + L & 53.5          & 78.7          & 36.5          & \textbf{89.3} & 53.5          \\ \hline
                             & + L         & 52.4          & \textbf{78.6} & 35.1          & 89.2          & 52.6          \\
\multirow{1}{1em}{\textcircled{2}} & + D + L     & 53.9          & 78.5          & 37.1          & \textbf{89.3} & 53.9          \\
                             & + J + L     & \textbf{54.1} & 78.2          & \textbf{37.6} & \textbf{89.3} & \textbf{54.5} \\ \hline
\end{tabular}
\label{table:data_aug}
\end{table}

\myPara{Impact of different data augmentations.}
As delineated in Section \ref{sec:mif}, we introduce a triad of data augmentation strategies, namely Random drop, Spatial jitter, and Random length. To investigate the efficacy of these methods, we conducted a comparative analysis using the performance metric HOTA. The results of this experiment are illustrated in Table \ref{table:data_aug} (\textcircled{1}). It is worth noting that using Random drop and Spatial jitter brings a HOTA improvement of 0.8 and 0.9 percentage points, respectively, compared to not using any data augmentation. Using both technologies at the same time brought a 1.1 percentage point improvement. However, when Random length is added to the augmentation process, a decrease in performance is observed. To verify this trend, we conducted a second set of experiments as detailed in Table \ref{table:data_aug} (\textcircled{2}), which confirmed that Random length did not contribute to any positive benefit. Based on these results, we decide to retain only the first two data augmentation techniques, namely Random drop and Spatial jitter.

\begin{table}[!t]
\centering
\caption{Impact of random drop probability $p_i$ during training.}
\setlength{\tabcolsep}{2.8mm}
\begin{tabular}{|l|ccccc|}
\hline
$p_i$ & HOTA↑         & DetA↑         & AssA↑         & MOTA↑         & IDF1↑ \\ \hline
0     & 53.5          & 78.6          & 36.5          & 89.3          & 53.5  \\
0.1   & \textbf{54.3} & 78.4          & \textbf{37.8} & 89.2          & 54.2  \\
0.2   & 54.0          & 78.6          & 37.3          & 89.2          & 54.1  \\
0.3   & 53.6          & \textbf{78.7} & 36.6          & \textbf{89.3} & 54.2  \\ \hline
\end{tabular}
\label{motion}
\label{table:drop_prob}
\end{table}

\myPara{Impact of random drop probability during training.}
The approach of randomly dropping a historical observation with a certain probability $p_i$ enables the generation of more complex motion patterns, such as rapid movements of an object. The impact of different values of $p_i$ on the tracking performance is investigated in \Cref{table:drop_prob}. The results indicate that as $p_i$ gradually increases from 0 to 0.3, the performance of the tracking method varies. Notably, the highest score on the HOTA metric is achieved when $p_i$ is set to 0.1. This finding suggests that a moderate level of dropping historical observations can enhance the tracking performance, while excessively high probabilities of dropping such observations may have a negative impact on the tracking quality.

\begin{table}[t]
\centering
\setlength{\tabcolsep}{2.8mm}
\caption{Comparison of different pooling types.}
\begin{tabular}{|l|ccccc|}
\hline
              & HOTA↑         & DetA↑         & AssA↑         & MOTA↑         & IDF1↑         \\ \hline
\textit{sum}  & 51.7          & 78.2          & 34.3          & 89.2          & 51.9          \\
\textit{last} & 53.8          & \textbf{78.7} & 37.0          & 89.2          & 54.0          \\
\textit{mean} & \textbf{54.6} & 78.6          & \textbf{38.1} & \textbf{89.2} & \textbf{54.6} \\ \hline
\end{tabular}
\label{table:pooling}
\end{table}

\myPara{Impact of different memory pooling types.}
The motion predictor proposed in this study utilizes the encoder of a Transformer to capture the temporal dynamics of an object, resulting in the generation of memory. To feed this memory to the regression head, a pooling operation is necessary. Specifically, we explore three different types of pooling operations, namely \textit{mean}, \textit{sum}, and \textit{last}, which respectively refer to taking the mean value of the encoder memory along the time dimension, summing the memory, or utilizing the last moment representation. 
The results presented in \Cref{table:pooling} demonstrate that the \textit{mean} pooling operation achieves the best performance among the three tested options. 


\begin{table}[!t]
\centering
\caption{The impact of maximum historical observation window.}
\setlength{\tabcolsep}{2.8mm}
\begin{tabular}{|l|ccccc|}
\hline
$n_{past}$ & HOTA↑  & DetA↑  & AssA↑  & MOTA↑  & IDF1↑  \\
\hline
3   & 51.6 & 78.2  & 34.2 & \textbf{89.2} & 51.2  \\
5   & 52.3 & 78.2  & 35.2 & \textbf{89.2} & 52.8 \\
10  & \textbf{54.6}  & \textbf{78.6}  & \textbf{38.1}  & \textbf{89.2}  & \textbf{54.6}  \\
13  & 54.2 & 78.3  & 37.8  & 89.1  & 54.1  \\
15  & 53.3 & 78.4 & 36.4 & 89.1 & 53.3 \\
\hline
\end{tabular}
\label{table:window_size.}
\end{table}

\myPara{Impact of maximum historical observation window.}
As presented in \cref{table:window_size.}, very small historical observation windows fail to provide sufficient information, resulting in inaccurate predictions. In contrast, very large historical observation windows introduce a considerable amount of noise, leading to degraded performance. Based on the results, we set the value of $n_{past}$ to 10, as it attains the best performance across all metrics. This finding highlights the importance of selecting an appropriate value for the historical observation window size to achieve optimal performance in object tracking tasks.




\section{Conclusion}
In this paper, we present MotionTrack, an innovative online tracker with a learnable motion predictor. By relying solely on the object's trajectory information, MotionTrack enhances robustness in challenging scenarios, including nonlinear motion and occlusion.
The proposed predictor incorporates two distinct modules to capture information at varying levels of granularity, thus enabling efficient modeling of an object's temporal dynamics. Specifically, we employ the Transformer encoder for motion prediction, which are capable of capturing complex motion patterns at the token level. We further introduce DyMLP, a MLP-like architecture, to extract semantic information distributed across different channels. By integrating DyMLP with the self-attention module in Transformer, our approach leverages the dual-granularity information to achieve superior performance. The results of our experiments have illustrated that the proposed methodology surpasses current state-of-the-art techniques when applied to datasets featuring intricate motion scenarios. Additionally, the performance of the proposed method has exhibited its resilience in the face of complex motion patterns.






\bibliographystyle{elsarticle-num-names} 
\bibliography{ref}

\begin{thebibliography}{71}
\expandafter\ifx\csname natexlab\endcsname\relax\def\natexlab#1{#1}\fi
\providecommand{\url}[1]{\texttt{#1}}
\providecommand{\href}[2]{#2}
\providecommand{\path}[1]{#1}
\providecommand{\DOIprefix}{doi:}
\providecommand{\ArXivprefix}{arXiv:}
\providecommand{\URLprefix}{URL: }
\providecommand{\Pubmedprefix}{pmid:}
\providecommand{\doi}[1]{\href{http://dx.doi.org/#1}{\path{#1}}}
\providecommand{\Pubmed}[1]{\href{pmid:#1}{\path{#1}}}
\providecommand{\bibinfo}[2]{#2}
\ifx\xfnm\relax \def\xfnm[#1]{\unskip,\space#1}\fi
\bibitem[{Wen et~al.(2020)Wen, Du, Cai, Lei, Chang, Qi, Lim, Yang, and Lyu}]{wen2020ua}
\bibinfo{author}{L.~Wen}, \bibinfo{author}{D.~Du}, \bibinfo{author}{Z.~Cai}, \bibinfo{author}{Z.~Lei}, \bibinfo{author}{M.-C. Chang}, \bibinfo{author}{H.~Qi}, \bibinfo{author}{J.~Lim}, \bibinfo{author}{M.-H. Yang}, \bibinfo{author}{S.~Lyu},
\newblock \bibinfo{title}{Ua-detrac: A new benchmark and protocol for multi-object detection and tracking},
\newblock \bibinfo{journal}{Computer Vision and Image Understanding} \bibinfo{volume}{193} (\bibinfo{year}{2020}) \bibinfo{pages}{102907}.
\bibitem[{Yuan et~al.(2022)Yuan, Iqbal, Molchanov, Kitani, and Kautz}]{glamr}
\bibinfo{author}{Y.~Yuan}, \bibinfo{author}{U.~Iqbal}, \bibinfo{author}{P.~Molchanov}, \bibinfo{author}{K.~Kitani}, \bibinfo{author}{J.~Kautz},
\newblock \bibinfo{title}{Glamr: Global occlusion-aware human mesh recovery with dynamic cameras},
\newblock in: \bibinfo{booktitle}{Proceedings of the IEEE/CVF Conference on Computer Vision and Pattern Recognition}, \bibinfo{year}{2022}, pp. \bibinfo{pages}{11038--11049}.
\bibitem[{Caesar et~al.(2020)Caesar, Bankiti, Lang, Vora, Liong, Xu, Krishnan, Pan, Baldan, and Beijbom}]{nuscenes}
\bibinfo{author}{H.~Caesar}, \bibinfo{author}{V.~Bankiti}, \bibinfo{author}{A.~H. Lang}, \bibinfo{author}{S.~Vora}, \bibinfo{author}{V.~E. Liong}, \bibinfo{author}{Q.~Xu}, \bibinfo{author}{A.~Krishnan}, \bibinfo{author}{Y.~Pan}, \bibinfo{author}{G.~Baldan}, \bibinfo{author}{O.~Beijbom},
\newblock \bibinfo{title}{nuscenes: A multimodal dataset for autonomous driving},
\newblock in: \bibinfo{booktitle}{Proceedings of the IEEE/CVF conference on computer vision and pattern recognition}, \bibinfo{year}{2020}, pp. \bibinfo{pages}{11621--11631}.
\bibitem[{Sun et~al.(2020)Sun, Kretzschmar, Dotiwalla, Chouard, Patnaik, Tsui, Guo, Zhou, Chai, Caine et~al.}]{sun2020scalability}
\bibinfo{author}{P.~Sun}, \bibinfo{author}{H.~Kretzschmar}, \bibinfo{author}{X.~Dotiwalla}, \bibinfo{author}{A.~Chouard}, \bibinfo{author}{V.~Patnaik}, \bibinfo{author}{P.~Tsui}, \bibinfo{author}{J.~Guo}, \bibinfo{author}{Y.~Zhou}, \bibinfo{author}{Y.~Chai}, \bibinfo{author}{B.~Caine}, et~al.,
\newblock \bibinfo{title}{Scalability in perception for autonomous driving: Waymo open dataset},
\newblock in: \bibinfo{booktitle}{Proceedings of the IEEE/CVF conference on computer vision and pattern recognition}, \bibinfo{year}{2020}, pp. \bibinfo{pages}{2446--2454}.
\bibitem[{Martin-Martin et~al.(2021)Martin-Martin, Patel, Rezatofighi, Shenoi, Gwak, Frankel, Sadeghian, and Savarese}]{jrdb}
\bibinfo{author}{R.~Martin-Martin}, \bibinfo{author}{M.~Patel}, \bibinfo{author}{H.~Rezatofighi}, \bibinfo{author}{A.~Shenoi}, \bibinfo{author}{J.~Gwak}, \bibinfo{author}{E.~Frankel}, \bibinfo{author}{A.~Sadeghian}, \bibinfo{author}{S.~Savarese},
\newblock \bibinfo{title}{Jrdb: A dataset and benchmark of egocentric robot visual perception of humans in built environments},
\newblock \bibinfo{journal}{IEEE transactions on pattern analysis and machine intelligence}  (\bibinfo{year}{2021}).
\bibitem[{Ren et~al.(2015)Ren, He, Girshick, and Sun}]{fasterrcnn}
\bibinfo{author}{S.~Ren}, \bibinfo{author}{K.~He}, \bibinfo{author}{R.~Girshick}, \bibinfo{author}{J.~Sun},
\newblock \bibinfo{title}{Faster r-cnn: Towards real-time object detection with region proposal networks},
\newblock \bibinfo{journal}{Advances in neural information processing systems} \bibinfo{volume}{28} (\bibinfo{year}{2015}).
\bibitem[{Redmon and Farhadi(2018)}]{yolov3}
\bibinfo{author}{J.~Redmon}, \bibinfo{author}{A.~Farhadi},
\newblock \bibinfo{title}{Yolov3: An incremental improvement},
\newblock \bibinfo{journal}{arXiv preprint arXiv:1804.02767}  (\bibinfo{year}{2018}).
\bibitem[{Zhou et~al.(2019)Zhou, Wang, and Kr{\"a}henb{\"u}hl}]{centernet}
\bibinfo{author}{X.~Zhou}, \bibinfo{author}{D.~Wang}, \bibinfo{author}{P.~Kr{\"a}henb{\"u}hl},
\newblock \bibinfo{title}{Objects as points},
\newblock in: \bibinfo{booktitle}{arXiv preprint arXiv:1904.07850}, \bibinfo{year}{2019}.
\bibitem[{Ge et~al.(2021)Ge, Liu, Wang, Li, and Sun}]{yolox}
\bibinfo{author}{Z.~Ge}, \bibinfo{author}{S.~Liu}, \bibinfo{author}{F.~Wang}, \bibinfo{author}{Z.~Li}, \bibinfo{author}{J.~Sun},
\newblock \bibinfo{title}{Yolox: Exceeding yolo series in 2021},
\newblock \bibinfo{journal}{arXiv preprint arXiv:2107.08430}  (\bibinfo{year}{2021}).
\bibitem[{Yang et~al.(2019)Yang, Li, and Dou}]{yang2019towards}
\bibinfo{author}{K.~Yang}, \bibinfo{author}{D.~Li}, \bibinfo{author}{Y.~Dou},
\newblock \bibinfo{title}{Towards precise end-to-end weakly supervised object detection network},
\newblock in: \bibinfo{booktitle}{Proceedings of the IEEE/CVF International Conference on Computer Vision}, \bibinfo{year}{2019}, pp. \bibinfo{pages}{8372--8381}.
\bibitem[{Wojke et~al.(2017)Wojke, Bewley, and Paulus}]{deepsort}
\bibinfo{author}{N.~Wojke}, \bibinfo{author}{A.~Bewley}, \bibinfo{author}{D.~Paulus},
\newblock \bibinfo{title}{Simple online and realtime tracking with a deep association metric},
\newblock in: \bibinfo{booktitle}{2017 IEEE international conference on image processing (ICIP)}, \bibinfo{organization}{IEEE}, \bibinfo{year}{2017}, pp. \bibinfo{pages}{3645--3649}.
\bibitem[{Hermans et~al.(2017)Hermans, Beyer, and Leibe}]{hermans2017defense}
\bibinfo{author}{A.~Hermans}, \bibinfo{author}{L.~Beyer}, \bibinfo{author}{B.~Leibe},
\newblock \bibinfo{title}{In defense of the triplet loss for person re-identification},
\newblock \bibinfo{journal}{arXiv preprint arXiv:1703.07737}  (\bibinfo{year}{2017}).
\bibitem[{Wang et~al.(2020)Wang, Zheng, Liu, Li, and Wang}]{jde}
\bibinfo{author}{Z.~Wang}, \bibinfo{author}{L.~Zheng}, \bibinfo{author}{Y.~Liu}, \bibinfo{author}{Y.~Li}, \bibinfo{author}{S.~Wang},
\newblock \bibinfo{title}{Towards real-time multi-object tracking},
\newblock in: \bibinfo{booktitle}{European Conference on Computer Vision}, \bibinfo{organization}{Springer}, \bibinfo{year}{2020}, pp. \bibinfo{pages}{107--122}.
\bibitem[{Zhang et~al.(2021)Zhang, Wang, Wang, Zeng, and Liu}]{fairmot}
\bibinfo{author}{Y.~Zhang}, \bibinfo{author}{C.~Wang}, \bibinfo{author}{X.~Wang}, \bibinfo{author}{W.~Zeng}, \bibinfo{author}{W.~Liu},
\newblock \bibinfo{title}{Fairmot: On the fairness of detection and re-identification in multiple object tracking},
\newblock \bibinfo{journal}{International Journal of Computer Vision} \bibinfo{volume}{129} (\bibinfo{year}{2021}) \bibinfo{pages}{3069--3087}.
\bibitem[{Wang et~al.(2021)Wang, Luo, and Zhu}]{wang2021two}
\bibinfo{author}{F.~Wang}, \bibinfo{author}{L.~Luo}, \bibinfo{author}{E.~Zhu},
\newblock \bibinfo{title}{Two-stage real-time multi-object tracking with candidate selection},
\newblock in: \bibinfo{booktitle}{International Conference on Multimedia Modeling}, \bibinfo{organization}{Springer}, \bibinfo{year}{2021}, pp. \bibinfo{pages}{49--61}.
\bibitem[{Milan et~al.(2016)Milan, Leal-Taix{\'e}, Reid, Roth, and Schindler}]{mot16}
\bibinfo{author}{A.~Milan}, \bibinfo{author}{L.~Leal-Taix{\'e}}, \bibinfo{author}{I.~Reid}, \bibinfo{author}{S.~Roth}, \bibinfo{author}{K.~Schindler},
\newblock \bibinfo{title}{Mot16: A benchmark for multi-object tracking},
\newblock \bibinfo{journal}{arXiv preprint arXiv:1603.00831}  (\bibinfo{year}{2016}).
\bibitem[{Dendorfer et~al.(2020)Dendorfer, Rezatofighi, Milan, Shi, Cremers, Reid, Roth, Schindler, and Leal-Taix{\'e}}]{mot20}
\bibinfo{author}{P.~Dendorfer}, \bibinfo{author}{H.~Rezatofighi}, \bibinfo{author}{A.~Milan}, \bibinfo{author}{J.~Shi}, \bibinfo{author}{D.~Cremers}, \bibinfo{author}{I.~Reid}, \bibinfo{author}{S.~Roth}, \bibinfo{author}{K.~Schindler}, \bibinfo{author}{L.~Leal-Taix{\'e}},
\newblock \bibinfo{title}{Mot20: A benchmark for multi object tracking in crowded scenes},
\newblock \bibinfo{journal}{arXiv preprint arXiv:2003.09003}  (\bibinfo{year}{2020}).
\bibitem[{Meinhardt et~al.(2022)Meinhardt, Kirillov, Leal-Taixe, and Feichtenhofer}]{trackformer}
\bibinfo{author}{T.~Meinhardt}, \bibinfo{author}{A.~Kirillov}, \bibinfo{author}{L.~Leal-Taixe}, \bibinfo{author}{C.~Feichtenhofer},
\newblock \bibinfo{title}{Trackformer: Multi-object tracking with transformers},
\newblock in: \bibinfo{booktitle}{The IEEE Conference on Computer Vision and Pattern Recognition (CVPR)}, \bibinfo{year}{2022}.
\bibitem[{Cui et~al.(2023)Cui, Zeng, Zhao, Yang, Wu, and Wang}]{SportsMOT}
\bibinfo{author}{Y.~Cui}, \bibinfo{author}{C.~Zeng}, \bibinfo{author}{X.~Zhao}, \bibinfo{author}{Y.~Yang}, \bibinfo{author}{G.~Wu}, \bibinfo{author}{L.~Wang},
\newblock \bibinfo{title}{Sportsmot: A large multi-object tracking dataset in multiple sports scenes},
\newblock in: \bibinfo{booktitle}{Proceedings of the IEEE/CVF International Conference on Computer Vision (ICCV)}, \bibinfo{year}{2023}, pp. \bibinfo{pages}{9921--9931}.
\bibitem[{Sun et~al.(2022)Sun, Cao, Jiang, Yuan, Bai, Kitani, and Luo}]{dancetrack}
\bibinfo{author}{P.~Sun}, \bibinfo{author}{J.~Cao}, \bibinfo{author}{Y.~Jiang}, \bibinfo{author}{Z.~Yuan}, \bibinfo{author}{S.~Bai}, \bibinfo{author}{K.~Kitani}, \bibinfo{author}{P.~Luo},
\newblock \bibinfo{title}{Dancetrack: Multi-object tracking in uniform appearance and diverse motion},
\newblock in: \bibinfo{booktitle}{Proceedings of the IEEE/CVF Conference on Computer Vision and Pattern Recognition}, \bibinfo{year}{2022}, pp. \bibinfo{pages}{20993--21002}.
\bibitem[{Bewley et~al.(2016)Bewley, Ge, Ott, Ramos, and Upcroft}]{sort}
\bibinfo{author}{A.~Bewley}, \bibinfo{author}{Z.~Ge}, \bibinfo{author}{L.~Ott}, \bibinfo{author}{F.~Ramos}, \bibinfo{author}{B.~Upcroft},
\newblock \bibinfo{title}{Simple online and realtime tracking},
\newblock in: \bibinfo{booktitle}{2016 IEEE international conference on image processing (ICIP)}, \bibinfo{organization}{IEEE}, \bibinfo{year}{2016}, pp. \bibinfo{pages}{3464--3468}.
\bibitem[{Long et~al.(2018)Long, Haizhou, Zijie, and Chong}]{motdt}
\bibinfo{author}{C.~Long}, \bibinfo{author}{A.~Haizhou}, \bibinfo{author}{Z.~Zijie}, \bibinfo{author}{S.~Chong},
\newblock \bibinfo{title}{Real-time multiple people tracking with deeply learned candidate selection and person re-identification},
\newblock in: \bibinfo{booktitle}{ICME}, \bibinfo{year}{2018}.
\bibitem[{Zhang et~al.(2022)Zhang, Sun, Jiang, Yu, Weng, Yuan, Luo, Liu, and Wang}]{bytetrack}
\bibinfo{author}{Y.~Zhang}, \bibinfo{author}{P.~Sun}, \bibinfo{author}{Y.~Jiang}, \bibinfo{author}{D.~Yu}, \bibinfo{author}{F.~Weng}, \bibinfo{author}{Z.~Yuan}, \bibinfo{author}{P.~Luo}, \bibinfo{author}{W.~Liu}, \bibinfo{author}{X.~Wang},
\newblock \bibinfo{title}{Bytetrack: Multi-object tracking by associating every detection box},
\newblock in: \bibinfo{booktitle}{European Conference on Computer Vision}, \bibinfo{organization}{Springer}, \bibinfo{year}{2022}, pp. \bibinfo{pages}{1--21}.
\bibitem[{Zhang et~al.(2020)Zhang, Zhou, Chang, Wan, Wang, Wu, and Huang}]{FFT}
\bibinfo{author}{J.~Zhang}, \bibinfo{author}{S.~Zhou}, \bibinfo{author}{X.~Chang}, \bibinfo{author}{F.~Wan}, \bibinfo{author}{J.~Wang}, \bibinfo{author}{Y.~Wu}, \bibinfo{author}{D.~Huang},
\newblock \bibinfo{title}{Multiple object tracking by flowing and fusing},
\newblock \bibinfo{journal}{arXiv preprint arXiv:2001.11180}  (\bibinfo{year}{2020}).
\bibitem[{Zhou et~al.(2020)Zhou, Koltun, and Kr{\"a}henb{\"u}hl}]{centertrack}
\bibinfo{author}{X.~Zhou}, \bibinfo{author}{V.~Koltun}, \bibinfo{author}{P.~Kr{\"a}henb{\"u}hl},
\newblock \bibinfo{title}{Tracking objects as points},
\newblock in: \bibinfo{booktitle}{European Conference on Computer Vision}, \bibinfo{organization}{Springer}, \bibinfo{year}{2020}, pp. \bibinfo{pages}{474--490}.
\bibitem[{Saleh et~al.(2021)Saleh, Aliakbarian, Rezatofighi, Salzmann, and Gould}]{artist}
\bibinfo{author}{F.~Saleh}, \bibinfo{author}{S.~Aliakbarian}, \bibinfo{author}{H.~Rezatofighi}, \bibinfo{author}{M.~Salzmann}, \bibinfo{author}{S.~Gould},
\newblock \bibinfo{title}{Probabilistic tracklet scoring and inpainting for multiple object tracking},
\newblock in: \bibinfo{booktitle}{Proceedings of the IEEE/CVF Conference on Computer Vision and Pattern Recognition}, \bibinfo{year}{2021}, pp. \bibinfo{pages}{14329--14339}.
\bibitem[{Bergmann et~al.(2019)Bergmann, Meinhardt, and Leal-Taixe}]{Tracktor}
\bibinfo{author}{P.~Bergmann}, \bibinfo{author}{T.~Meinhardt}, \bibinfo{author}{L.~Leal-Taixe},
\newblock \bibinfo{title}{Tracking without bells and whistles},
\newblock in: \bibinfo{booktitle}{Proceedings of the IEEE/CVF International Conference on Computer Vision}, \bibinfo{year}{2019}, pp. \bibinfo{pages}{941--951}.
\bibitem[{Kalman et~al.(1960)}]{kalman}
\bibinfo{author}{R.~E. Kalman}, et~al.,
\newblock \bibinfo{title}{Contributions to the theory of optimal control},
\newblock \bibinfo{journal}{Bol. soc. mat. mexicana} \bibinfo{volume}{5} (\bibinfo{year}{1960}) \bibinfo{pages}{102--119}.
\bibitem[{Chaabane et~al.(2021)Chaabane, Zhang, Beveridge, and O'Hara}]{deft}
\bibinfo{author}{M.~Chaabane}, \bibinfo{author}{P.~Zhang}, \bibinfo{author}{R.~Beveridge}, \bibinfo{author}{S.~O'Hara},
\newblock \bibinfo{title}{Deft: Detection embeddings for tracking},
\newblock \bibinfo{journal}{arXiv preprint arXiv:2102.02267}  (\bibinfo{year}{2021}).
\bibitem[{Luo et~al.(2018)Luo, Yang, and Urtasun}]{luo2018fast}
\bibinfo{author}{W.~Luo}, \bibinfo{author}{B.~Yang}, \bibinfo{author}{R.~Urtasun},
\newblock \bibinfo{title}{Fast and furious: Real time end-to-end 3d detection, tracking and motion forecasting with a single convolutional net},
\newblock in: \bibinfo{booktitle}{Proceedings of the IEEE conference on Computer Vision and Pattern Recognition}, \bibinfo{year}{2018}, pp. \bibinfo{pages}{3569--3577}.
\bibitem[{Bai et~al.(2018)Bai, Kolter, and Koltun}]{bai2018empirical}
\bibinfo{author}{S.~Bai}, \bibinfo{author}{J.~Z. Kolter}, \bibinfo{author}{V.~Koltun},
\newblock \bibinfo{title}{An empirical evaluation of generic convolutional and recurrent networks for sequence modeling},
\newblock \bibinfo{journal}{arXiv preprint arXiv:1803.01271}  (\bibinfo{year}{2018}).
\bibitem[{Cao et~al.(2023)Cao, Pang, Weng, Khirodkar, and Kitani}]{ocsort}
\bibinfo{author}{J.~Cao}, \bibinfo{author}{J.~Pang}, \bibinfo{author}{X.~Weng}, \bibinfo{author}{R.~Khirodkar}, \bibinfo{author}{K.~Kitani},
\newblock \bibinfo{title}{Observation-centric sort: Rethinking sort for robust multi-object tracking},
\newblock in: \bibinfo{booktitle}{Proceedings of the IEEE Conference on Computer Vision and Pattern Recognition}, \bibinfo{year}{2023}, pp. \bibinfo{pages}{9686--9696}.
\bibitem[{Zhou et~al.(2021)Zhou, Zhang, Peng, Zhang, Li, Xiong, and Zhang}]{zhou2021informer}
\bibinfo{author}{H.~Zhou}, \bibinfo{author}{S.~Zhang}, \bibinfo{author}{J.~Peng}, \bibinfo{author}{S.~Zhang}, \bibinfo{author}{J.~Li}, \bibinfo{author}{H.~Xiong}, \bibinfo{author}{W.~Zhang},
\newblock \bibinfo{title}{Informer: Beyond efficient transformer for long sequence time-series forecasting},
\newblock in: \bibinfo{booktitle}{Proceedings of the AAAI conference on artificial intelligence}, volume~\bibinfo{volume}{35}, \bibinfo{year}{2021}, pp. \bibinfo{pages}{11106--11115}.
\bibitem[{Vaswani et~al.(2017)Vaswani, Shazeer, Parmar, Uszkoreit, Jones, Gomez, Kaiser, and Polosukhin}]{transformer}
\bibinfo{author}{A.~Vaswani}, \bibinfo{author}{N.~Shazeer}, \bibinfo{author}{N.~Parmar}, \bibinfo{author}{J.~Uszkoreit}, \bibinfo{author}{L.~Jones}, \bibinfo{author}{A.~N. Gomez}, \bibinfo{author}{{\L}.~Kaiser}, \bibinfo{author}{I.~Polosukhin},
\newblock \bibinfo{title}{Attention is all you need},
\newblock \bibinfo{journal}{Advances in neural information processing systems} \bibinfo{volume}{30} (\bibinfo{year}{2017}).
\bibitem[{Bau et~al.(2020)Bau, Zhu, Strobelt, Lapedriza, Zhou, and Torralba}]{bau2020understanding}
\bibinfo{author}{D.~Bau}, \bibinfo{author}{J.-Y. Zhu}, \bibinfo{author}{H.~Strobelt}, \bibinfo{author}{A.~Lapedriza}, \bibinfo{author}{B.~Zhou}, \bibinfo{author}{A.~Torralba},
\newblock \bibinfo{title}{Understanding the role of individual units in a deep neural network},
\newblock \bibinfo{journal}{Proceedings of the National Academy of Sciences} \bibinfo{volume}{117} (\bibinfo{year}{2020}) \bibinfo{pages}{30071--30078}.
\bibitem[{Wu et~al.(2021)Wu, Lischinski, and Shechtman}]{wu2021stylespace}
\bibinfo{author}{Z.~Wu}, \bibinfo{author}{D.~Lischinski}, \bibinfo{author}{E.~Shechtman},
\newblock \bibinfo{title}{Stylespace analysis: Disentangled controls for stylegan image generation},
\newblock in: \bibinfo{booktitle}{Proceedings of the IEEE/CVF Conference on Computer Vision and Pattern Recognition}, \bibinfo{year}{2021}, pp. \bibinfo{pages}{12863--12872}.
\bibitem[{Li et~al.(2023)Li, Rao, Pan, and Xu}]{Li2023MTSMixersMT}
\bibinfo{author}{Z.~Li}, \bibinfo{author}{Z.~Rao}, \bibinfo{author}{L.~Pan}, \bibinfo{author}{Z.~Xu},
\newblock \bibinfo{title}{Mts-mixers: Multivariate time series forecasting via factorized temporal and channel mixing},
\newblock \bibinfo{journal}{ArXiv} \bibinfo{volume}{abs/2302.04501} (\bibinfo{year}{2023}).
\bibitem[{Zhang and Yan(2023)}]{zhang2023crossformer}
\bibinfo{author}{Y.~Zhang}, \bibinfo{author}{J.~Yan},
\newblock \bibinfo{title}{Crossformer: Transformer utilizing cross-dimension dependency for multivariate time series forecasting},
\newblock in: \bibinfo{booktitle}{The Eleventh International Conference on Learning Representations}, \bibinfo{year}{2023}.
\bibitem[{Chen et~al.(2023)Chen, Li, Arik, Yoder, and Pfister}]{chen2023tsmixer}
\bibinfo{author}{S.-A. Chen}, \bibinfo{author}{C.-L. Li}, \bibinfo{author}{S.~O. Arik}, \bibinfo{author}{N.~C. Yoder}, \bibinfo{author}{T.~Pfister},
\newblock \bibinfo{title}{{TSM}ixer: An all-{MLP} architecture for time series forecast-ing},
\newblock \bibinfo{journal}{Transactions on Machine Learning Research}  (\bibinfo{year}{2023}).
\bibitem[{Roshan~Zamir et~al.(2012)Roshan~Zamir, Dehghan, and Shah}]{roshan2012gmcp}
\bibinfo{author}{A.~Roshan~Zamir}, \bibinfo{author}{A.~Dehghan}, \bibinfo{author}{M.~Shah},
\newblock \bibinfo{title}{Gmcp-tracker: Global multi-object tracking using generalized minimum clique graphs},
\newblock in: \bibinfo{booktitle}{European conference on computer vision}, \bibinfo{organization}{Springer}, \bibinfo{year}{2012}, pp. \bibinfo{pages}{343--356}.
\bibitem[{Wen et~al.(2014)Wen, Li, Yan, Lei, Yi, and Li}]{wen2014}
\bibinfo{author}{L.~Wen}, \bibinfo{author}{W.~Li}, \bibinfo{author}{J.~Yan}, \bibinfo{author}{Z.~Lei}, \bibinfo{author}{D.~Yi}, \bibinfo{author}{S.~Z. Li},
\newblock \bibinfo{title}{Multiple target tracking based on undirected hierarchical relation hypergraph},
\newblock in: \bibinfo{booktitle}{Proceedings of the IEEE conference on computer vision and pattern recognition}, \bibinfo{year}{2014}, pp. \bibinfo{pages}{1282--1289}.
\bibitem[{Lan et~al.(2016)Lan, Tao, Gong, Guan, and Luo}]{lan}
\bibinfo{author}{L.~Lan}, \bibinfo{author}{D.~Tao}, \bibinfo{author}{C.~Gong}, \bibinfo{author}{N.~Guan}, \bibinfo{author}{Z.~Luo},
\newblock \bibinfo{title}{Online multi-object tracking by quadratic pseudo-boolean optimization.},
\newblock in: \bibinfo{booktitle}{IJCAI}, \bibinfo{year}{2016}, pp. \bibinfo{pages}{3396--3402}.
\bibitem[{Zhang and Yang(2021)}]{zhang2021survey}
\bibinfo{author}{Y.~Zhang}, \bibinfo{author}{Q.~Yang},
\newblock \bibinfo{title}{A survey on multi-task learning},
\newblock \bibinfo{journal}{IEEE Transactions on Knowledge and Data Engineering}  (\bibinfo{year}{2021}).
\bibitem[{Milan et~al.(2017)Milan, Rezatofighi, Dick, Reid, and Schindler}]{milan2017}
\bibinfo{author}{A.~Milan}, \bibinfo{author}{S.~H. Rezatofighi}, \bibinfo{author}{A.~Dick}, \bibinfo{author}{I.~Reid}, \bibinfo{author}{K.~Schindler},
\newblock \bibinfo{title}{Online multi-target tracking using recurrent neural networks},
\newblock in: \bibinfo{booktitle}{Thirty-First AAAI conference on artificial intelligence}, \bibinfo{year}{2017}.
\bibitem[{Wan et~al.(2018)Wan, Wang, and Zhou}]{rnn1}
\bibinfo{author}{X.~Wan}, \bibinfo{author}{J.~Wang}, \bibinfo{author}{S.~Zhou},
\newblock \bibinfo{title}{An online and flexible multi-object tracking framework using long short-term memory},
\newblock in: \bibinfo{booktitle}{Proceedings of the IEEE Conference on Computer Vision and Pattern Recognition Workshops}, \bibinfo{year}{2018}, pp. \bibinfo{pages}{1230--1238}.
\bibitem[{Sadeghian et~al.(2017)Sadeghian, Alahi, and Savarese}]{rnn2}
\bibinfo{author}{A.~Sadeghian}, \bibinfo{author}{A.~Alahi}, \bibinfo{author}{S.~Savarese},
\newblock \bibinfo{title}{Tracking the untrackable: Learning to track multiple cues with long-term dependencies},
\newblock in: \bibinfo{booktitle}{Proceedings of the IEEE international conference on computer vision}, \bibinfo{year}{2017}, pp. \bibinfo{pages}{300--311}.
\bibitem[{Ran et~al.(2019)Ran, Kong, Wang, and Liu}]{rnn3}
\bibinfo{author}{N.~Ran}, \bibinfo{author}{L.~Kong}, \bibinfo{author}{Y.~Wang}, \bibinfo{author}{Q.~Liu},
\newblock \bibinfo{title}{A robust multi-athlete tracking algorithm by exploiting discriminant features and long-term dependencies},
\newblock in: \bibinfo{booktitle}{International Conference on Multimedia Modeling}, \bibinfo{organization}{Springer}, \bibinfo{year}{2019}, pp. \bibinfo{pages}{411--423}.
\bibitem[{Stadler and Beyerer(2021)}]{TMOH}
\bibinfo{author}{D.~Stadler}, \bibinfo{author}{J.~Beyerer},
\newblock \bibinfo{title}{Improving multiple pedestrian tracking by track management and occlusion handling},
\newblock in: \bibinfo{booktitle}{Proceedings of the IEEE/CVF conference on computer vision and pattern recognition}, \bibinfo{year}{2021}, pp. \bibinfo{pages}{10958--10967}.
\bibitem[{Han et~al.(2022)Han, Huang, Wang, Yu, Liu, and Pan}]{mat}
\bibinfo{author}{S.~Han}, \bibinfo{author}{P.~Huang}, \bibinfo{author}{H.~Wang}, \bibinfo{author}{E.~Yu}, \bibinfo{author}{D.~Liu}, \bibinfo{author}{X.~Pan},
\newblock \bibinfo{title}{Mat: Motion-aware multi-object tracking},
\newblock \bibinfo{journal}{Neurocomputing} \bibinfo{volume}{476} (\bibinfo{year}{2022}) \bibinfo{pages}{75--86}.
\bibitem[{Dendorfer et~al.(2022)Dendorfer, Yugay, O{\v{s}}ep, and Leal-Taix{\'e}}]{QuoVadis}
\bibinfo{author}{P.~Dendorfer}, \bibinfo{author}{V.~Yugay}, \bibinfo{author}{A.~O{\v{s}}ep}, \bibinfo{author}{L.~Leal-Taix{\'e}},
\newblock \bibinfo{title}{Quo vadis: Is trajectory forecasting the key towards long-term multi-object tracking?},
\newblock \bibinfo{journal}{Advances in neural information processing systems}  (\bibinfo{year}{2022}).
\bibitem[{Carion et~al.(2020)Carion, Massa, Synnaeve, Usunier, Kirillov, and Zagoruyko}]{detr}
\bibinfo{author}{N.~Carion}, \bibinfo{author}{F.~Massa}, \bibinfo{author}{G.~Synnaeve}, \bibinfo{author}{N.~Usunier}, \bibinfo{author}{A.~Kirillov}, \bibinfo{author}{S.~Zagoruyko},
\newblock \bibinfo{title}{End-to-end object detection with transformers},
\newblock in: \bibinfo{booktitle}{European conference on computer vision}, \bibinfo{organization}{Springer}, \bibinfo{year}{2020}, pp. \bibinfo{pages}{213--229}.
\bibitem[{Zhu et~al.(2020)Zhu, Su, Lu, Li, Wang, and Dai}]{defdetr}
\bibinfo{author}{X.~Zhu}, \bibinfo{author}{W.~Su}, \bibinfo{author}{L.~Lu}, \bibinfo{author}{B.~Li}, \bibinfo{author}{X.~Wang}, \bibinfo{author}{J.~Dai},
\newblock \bibinfo{title}{Deformable detr: Deformable transformers for end-to-end object detection},
\newblock \bibinfo{journal}{ICLR}  (\bibinfo{year}{2020}).
\bibitem[{Wang et~al.(2024)Wang, Lai, Wang, and Zhang}]{WANG2024106110}
\bibinfo{author}{J.~Wang}, \bibinfo{author}{C.~Lai}, \bibinfo{author}{Y.~Wang}, \bibinfo{author}{W.~Zhang},
\newblock \bibinfo{title}{Emat: Efficient feature fusion network for visual tracking via optimized multi-head attention},
\newblock \bibinfo{journal}{Neural Networks} \bibinfo{volume}{172} (\bibinfo{year}{2024}) \bibinfo{pages}{106110}.
\bibitem[{Cai et~al.(2024)Cai, Lan, Zhang, Zhang, Zhan, and Luo}]{CAI2024548}
\bibinfo{author}{H.~Cai}, \bibinfo{author}{L.~Lan}, \bibinfo{author}{J.~Zhang}, \bibinfo{author}{X.~Zhang}, \bibinfo{author}{Y.~Zhan}, \bibinfo{author}{Z.~Luo},
\newblock \bibinfo{title}{Iouformer: Pseudo-iou prediction with transformer for visual tracking},
\newblock \bibinfo{journal}{Neural Networks} \bibinfo{volume}{170} (\bibinfo{year}{2024}) \bibinfo{pages}{548--563}.
\bibitem[{Sun et~al.(2020)Sun, Cao, Jiang, Zhang, Xie, Yuan, Wang, and Luo}]{transtrack}
\bibinfo{author}{P.~Sun}, \bibinfo{author}{J.~Cao}, \bibinfo{author}{Y.~Jiang}, \bibinfo{author}{R.~Zhang}, \bibinfo{author}{E.~Xie}, \bibinfo{author}{Z.~Yuan}, \bibinfo{author}{C.~Wang}, \bibinfo{author}{P.~Luo},
\newblock \bibinfo{title}{Transtrack: Multiple object tracking with transformer},
\newblock \bibinfo{journal}{arXiv preprint arXiv:2012.15460}  (\bibinfo{year}{2020}).
\bibitem[{Xu et~al.(2021)Xu, Ban, Delorme, Gan, Rus, and Alameda-Pineda}]{transcenter}
\bibinfo{author}{Y.~Xu}, \bibinfo{author}{Y.~Ban}, \bibinfo{author}{G.~Delorme}, \bibinfo{author}{C.~Gan}, \bibinfo{author}{D.~Rus}, \bibinfo{author}{X.~Alameda-Pineda}, \bibinfo{title}{Transcenter: Transformers with dense representations for multiple-object tracking}, \bibinfo{year}{2021}. \href{http://arxiv.org/abs/2103.15145}{{\tt arXiv:2103.15145}}.
\bibitem[{Zeng et~al.(2022)Zeng, Dong, Zhang, Wang, Zhang, and Wei}]{motr}
\bibinfo{author}{F.~Zeng}, \bibinfo{author}{B.~Dong}, \bibinfo{author}{Y.~Zhang}, \bibinfo{author}{T.~Wang}, \bibinfo{author}{X.~Zhang}, \bibinfo{author}{Y.~Wei},
\newblock \bibinfo{title}{Motr: End-to-end multiple-object tracking with transformer},
\newblock in: \bibinfo{booktitle}{European Conference on Computer Vision (ECCV)}, \bibinfo{year}{2022}.
\bibitem[{Zhou et~al.(2022)Zhou, Yin, Koltun, and Kr{\"a}henb{\"u}hl}]{gtr}
\bibinfo{author}{X.~Zhou}, \bibinfo{author}{T.~Yin}, \bibinfo{author}{V.~Koltun}, \bibinfo{author}{P.~Kr{\"a}henb{\"u}hl},
\newblock \bibinfo{title}{Global tracking transformers},
\newblock in: \bibinfo{booktitle}{Proceedings of the IEEE/CVF Conference on Computer Vision and Pattern Recognition}, \bibinfo{year}{2022}, pp. \bibinfo{pages}{8771--8780}.
\bibitem[{Yu et~al.(2022)Yu, Li, Han, and Wang}]{relationtrack}
\bibinfo{author}{E.~Yu}, \bibinfo{author}{Z.~Li}, \bibinfo{author}{S.~Han}, \bibinfo{author}{H.~Wang},
\newblock \bibinfo{title}{Relationtrack: Relation-aware multiple object tracking with decoupled representation},
\newblock \bibinfo{journal}{IEEE Transactions on Multimedia}  (\bibinfo{year}{2022}).
\bibitem[{Bras{\'o} and Leal-Taix{\'e}(2020)}]{braso2020learning}
\bibinfo{author}{G.~Bras{\'o}}, \bibinfo{author}{L.~Leal-Taix{\'e}},
\newblock \bibinfo{title}{Learning a neural solver for multiple object tracking},
\newblock in: \bibinfo{booktitle}{Proceedings of the IEEE/CVF conference on computer vision and pattern recognition}, \bibinfo{year}{2020}, pp. \bibinfo{pages}{6247--6257}.
\bibitem[{Yuan et~al.(2021)Yuan, Weng, Ou, and Kitani}]{yuan2021agentformer}
\bibinfo{author}{Y.~Yuan}, \bibinfo{author}{X.~Weng}, \bibinfo{author}{Y.~Ou}, \bibinfo{author}{K.~M. Kitani},
\newblock \bibinfo{title}{Agentformer: Agent-aware transformers for socio-temporal multi-agent forecasting},
\newblock in: \bibinfo{booktitle}{Proceedings of the IEEE/CVF International Conference on Computer Vision}, \bibinfo{year}{2021}, pp. \bibinfo{pages}{9813--9823}.
\bibitem[{He et~al.(2016)He, Zhang, Ren, and Sun}]{he2016deep}
\bibinfo{author}{K.~He}, \bibinfo{author}{X.~Zhang}, \bibinfo{author}{S.~Ren}, \bibinfo{author}{J.~Sun},
\newblock \bibinfo{title}{Deep residual learning for image recognition},
\newblock in: \bibinfo{booktitle}{Proceedings of the IEEE conference on computer vision and pattern recognition}, \bibinfo{year}{2016}, pp. \bibinfo{pages}{770--778}.
\bibitem[{Ba et~al.(2016)Ba, Kiros, and Hinton}]{ba2016layer}
\bibinfo{author}{J.~L. Ba}, \bibinfo{author}{J.~R. Kiros}, \bibinfo{author}{G.~E. Hinton},
\newblock \bibinfo{title}{Layer normalization},
\newblock \bibinfo{journal}{arXiv preprint arXiv:1607.06450}  (\bibinfo{year}{2016}).
\bibitem[{Girshick(2015)}]{fastrcnn}
\bibinfo{author}{R.~Girshick},
\newblock \bibinfo{title}{Fast r-cnn},
\newblock in: \bibinfo{booktitle}{Proceedings of the IEEE international conference on computer vision}, \bibinfo{year}{2015}, pp. \bibinfo{pages}{1440--1448}.
\bibitem[{Kuhn(1955)}]{hungarian}
\bibinfo{author}{H.~W. Kuhn},
\newblock \bibinfo{title}{The hungarian method for the assignment problem},
\newblock \bibinfo{journal}{Naval research logistics quarterly} \bibinfo{volume}{2} (\bibinfo{year}{1955}) \bibinfo{pages}{83--97}.
\bibitem[{Luiten et~al.(2021)Luiten, Osep, Dendorfer, Torr, Geiger, Leal-Taix{\'e}, and Leibe}]{hota}
\bibinfo{author}{J.~Luiten}, \bibinfo{author}{A.~Osep}, \bibinfo{author}{P.~Dendorfer}, \bibinfo{author}{P.~Torr}, \bibinfo{author}{A.~Geiger}, \bibinfo{author}{L.~Leal-Taix{\'e}}, \bibinfo{author}{B.~Leibe},
\newblock \bibinfo{title}{Hota: A higher order metric for evaluating multi-object tracking},
\newblock \bibinfo{journal}{International journal of computer vision} \bibinfo{volume}{129} (\bibinfo{year}{2021}) \bibinfo{pages}{548--578}.
\bibitem[{Ristani and Solera(2016)}]{idf1}
\bibinfo{author}{E.~Ristani}, \bibinfo{author}{Solera},
\newblock \bibinfo{title}{Performance measures and a data set for multi-target, multi-camera tracking},
\newblock in: \bibinfo{booktitle}{European conference on computer vision}, \bibinfo{organization}{Springer}, \bibinfo{year}{2016}, pp. \bibinfo{pages}{17--35}.
\bibitem[{Ristani et~al.(2016)Ristani, Solera, Zou, Cucchiara, and Tomasi}]{clear}
\bibinfo{author}{E.~Ristani}, \bibinfo{author}{F.~Solera}, \bibinfo{author}{R.~Zou}, \bibinfo{author}{R.~Cucchiara}, \bibinfo{author}{C.~Tomasi},
\newblock \bibinfo{title}{Performance measures and a data set for multi-target, multi-camera tracking},
\newblock in: \bibinfo{booktitle}{Computer Vision--ECCV 2016 Workshops: Amsterdam, The Netherlands, October 8-10 and 15-16, 2016, Proceedings, Part II}, \bibinfo{organization}{Springer}, \bibinfo{year}{2016}, pp. \bibinfo{pages}{17--35}.
\bibitem[{Kingma and Ba(2014)}]{adam}
\bibinfo{author}{D.~P. Kingma}, \bibinfo{author}{J.~Ba},
\newblock \bibinfo{title}{Adam: A method for stochastic optimization},
\newblock \bibinfo{journal}{arXiv preprint arXiv:1412.6980}  (\bibinfo{year}{2014}).
\bibitem[{Pang et~al.(2021)Pang, Qiu, Li, Chen, Li, Darrell, and Yu}]{quasi}
\bibinfo{author}{J.~Pang}, \bibinfo{author}{L.~Qiu}, \bibinfo{author}{X.~Li}, \bibinfo{author}{H.~Chen}, \bibinfo{author}{Q.~Li}, \bibinfo{author}{T.~Darrell}, \bibinfo{author}{F.~Yu},
\newblock \bibinfo{title}{Quasi-dense similarity learning for multiple object tracking},
\newblock in: \bibinfo{booktitle}{Proceedings of the IEEE/CVF conference on computer vision and pattern recognition}, \bibinfo{year}{2021}, pp. \bibinfo{pages}{164--173}.
\bibitem[{Wu et~al.(2021)Wu, Cao, Song, Wang, Yang, and Yuan}]{wu2021track}
\bibinfo{author}{J.~Wu}, \bibinfo{author}{J.~Cao}, \bibinfo{author}{L.~Song}, \bibinfo{author}{Y.~Wang}, \bibinfo{author}{M.~Yang}, \bibinfo{author}{J.~Yuan},
\newblock \bibinfo{title}{Track to detect and segment: An online multi-object tracker},
\newblock in: \bibinfo{booktitle}{Proceedings of the IEEE/CVF conference on computer vision and pattern recognition}, \bibinfo{year}{2021}, pp. \bibinfo{pages}{12352--12361}.

\end{thebibliography}




\end{document}